\documentclass[english,ams,yap]{iitpinfo}
\usepackage[utf8]{inputenc}
\usepackage{hyperref}
\usepackage{amsmath}
\usepackage{array}
\usepackage{booktabs}
\usepackage{amssymb,xcolor,stackengine,graphicx}
\usepackage{multirow}
\usepackage{adjustbox}

\newcommand{\vecX}{\mathbf{x}}

\newcommand{\vecZ}{\mathbf{z}}
\newcommand{\vecU}{\mathbf{u}}
\newcommand{\vecM}{\boldsymbol{\mu}}

\newcommand{\calN}{\mathcal{N}}
\newcommand{\sphere}{\mathbb{S}^{d-1}}

\newcommand{\KL}{D_{\mathrm{KL}}}
\newcommand{\R}{\mathbb{R}}
\DeclareMathOperator*{\argmax}{argmax}

\VolumeNo{25}
\IssueNo{1}
\YearOfIssue{2025}
\CopyrightYear{2025}
\CopyrightedAuthors{Erlygin}

\makeatletter
\renewcommand{\inst}[1]{%
  \unskip
  $^{\@instsymbollist{#1}}$%
}
\newcommand{\@instsymbollist}[1]{%
  \@for\next:=#1\do{%
    \@instsymbol{\next}%
    \ifx\next\@lastofloop\else\,\fi
  }%
}
\makeatother

\title{Mixed-Prior Decision Risk for Calibrated Uncertainty Estimation in Open-Set Recognition\thanks{The research was supported by the Russian Science Foundation grant No. 25-11-00355}}

\author{L.\,A.\,Erlygin\inst{1}, P.\,D.\,Proskura\inst{3}, A.\,A.\,Zaytsev\inst{1} \inst{2}}

\institute{Skolkovo Institute of Science and Technology (Skoltech), Moscow, Russia \and Risk Management, Sber, Moscow, Russia \and The Institute for Information Transmission Problems (IITP RAS), Moscow, Russia}

\received{Received September 15, 2025}

\titlerunning{Mixed-Prior Decision Risk for Open-Set Recognition}

\authorrunning{ERLYGIN et al.}

\Rubric{MACHINE LEARNING AND PATTERN RECOGNITION}

\begin{document}

\maketitle

\begin{abstract}
In open-set recognition (OSR), a probe must either be identified as one of the known gallery classes or rejected as unknown, so three error types coexist: false acceptance, false rejection, and misidentification.
An uncertainty score for selective recognition should rank probes by the risk of the decision the system has made.

Bayesian gallery-aware models such as Holistic Uncertainty Estimation (HolUE) summarize the posterior over known and unknown classes by Kullback--Leibler (KL) divergence components and map them to an uncertainty score with a supervised nonlinear calibrator.
We show that the KL summary is not generally monotone in decision risk: linear fusion of the KL components tuned on validation data yields negative filtering quality on several benchmarks.

We propose \emph{MPRisk}, a mixed-prior posterior decision-risk score that keeps the same Bayesian posterior but directly scores the error events associated with the selected decision: false-acceptance, misidentification, and false-rejection risks, plus a non-specificity penalty for rejections, enabled by modeling unknown identities as a continuous component.
Four nonnegative weights tuned on a validation set suffice for ranking; no nonlinear supervised model is required.

Across nine image, audio, and text benchmarks, MPRisk achieves the best or tied-best Prediction Rejection Ratio at every operating point on the image and audio benchmarks and on most text operating points, with bootstrap-confirmed gains over HolUE on five benchmarks (up to $+0.19$ PRR) at comparable or lower runtime.
\end{abstract}

\textbf{KEYWORDS:} machine learning, open-set recognition, uncertainty estimation, selective prediction, Bayesian decision theory, probabilistic embeddings, risk decomposition.

\section{INTRODUCTION}\label{sec:intro}

Open-set recognition (OSR) is the standard formulation for recognition systems deployed in open-world environments: the system must both identify samples from known classes and reject samples from unknown classes.
This setting appears in face recognition, speaker identification, whale and dolphin identification, authorship attribution, intent classification, and topic recognition~\cite{scheirer2013toward,face_handbook,voxblink,cheeseman2022happywhale,clinc150,yahoo_ostc}.
A standard OSR pipeline compares the probe embedding to the gallery prototypes by cosine similarity: the probe is accepted if the best similarity exceeds a threshold and is then assigned the identity of the closest prototype; otherwise it is rejected as unknown.
With modern metric-learning losses such as ArcFace and CosFace, this paradigm achieves high recognition accuracy~\cite{deng2019arcface,wang2018cosface}.

High accuracy alone is not sufficient in risk-sensitive applications.
The system must also estimate the reliability of each decision, so that unreliable decisions can be deferred to a human operator or resolved by acquiring another sample.
This is the selective-recognition setting, where an uncertainty score is evaluated by how well it ranks erroneous decisions ahead of correct ones~\cite{1054406,NIPS2017_4a8423d5}.
In OSR the ranking task is non-trivial because three qualitatively different errors coexist:
\begin{enumerate}
    \item \textbf{False acceptance (FA):} an unknown probe is accepted as a known gallery class.
    \item \textbf{False rejection (FR):} a known probe is rejected as unknown.
    \item \textbf{Misidentification (ID):} a known probe is accepted but assigned the wrong identity.
\end{enumerate}

Gallery-aware Bayesian models address this multiplicity by reconstructing a posterior distribution over known and unknown classes.
Probabilistic embeddings such as PFE and SCF capture sample-quality ambiguity~\cite{pfe,scf}; gallery-aware models additionally use the relative position of the probe and the gallery prototypes~\cite{my_paper}.
Holistic Uncertainty Estimation (HolUE) combines both sources: it summarizes the posterior by two Kullback--Leibler (KL) divergence components and maps them to an uncertainty score with a supervised nonlinear calibrator trained on validation data.

The starting point of this paper is a general question: \emph{are information-gain summaries appropriate for ranking the risk of an OSR decision?}
KL divergence measures how far the posterior moved from the prior, whereas selective recognition requires a score aligned with the probability or cost of the actual decision error.
These two orderings can disagree: a posterior split between two plausible gallery identities is far from a uniform prior, yet the maximum-posterior decision carries high risk.
Section~\ref{sec:whykl} makes this mismatch precise, and Section~\ref{sec:fair} confirms it empirically: linear fusion of the HolUE KL components, tuned on validation data, produces negative filtering quality on several benchmarks, while the same features perform well under HolUE's nonlinear calibration.

\begin{figure*}[t]
\centering
\includegraphics[width=1\linewidth]{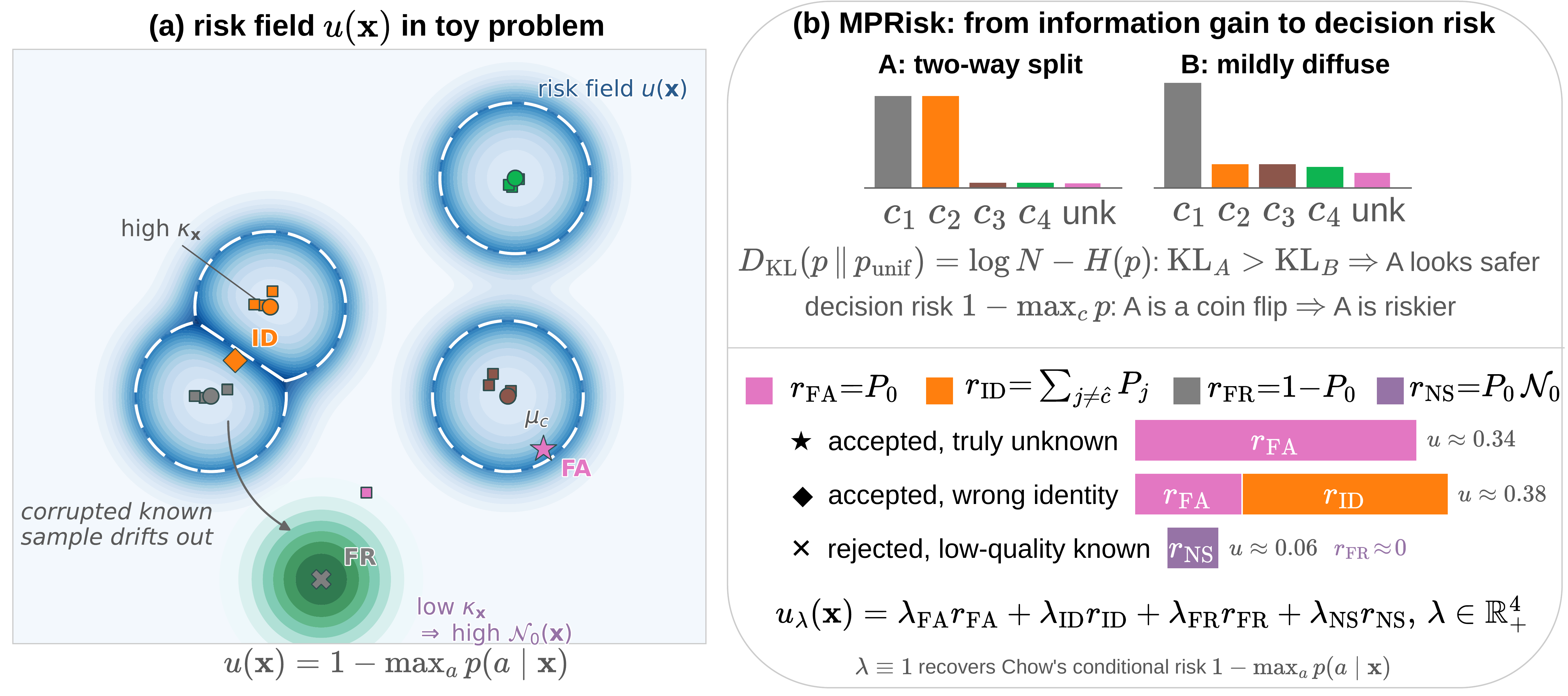}
\caption{MPRisk overview.
\textbf{(a)}~Embedding space of a fixed OSR system with $K$ gallery classes (prototypes $\bullet$, samples $\blacksquare$) under the mixed prior: unknown identities form a continuous component $c\in(K,K{+}1]$ (tinted reject region); dashed lines are decision boundaries; blue shading shows the decision-risk field $1-\max_a p(a\mid\vecX)$.
Markers show the three error types: false acceptance ($\bigstar$), misidentification ($\blacklozenge$), and false rejection ($\boldsymbol{\times}$) --- a corrupted known probe whose diffuse embedding distribution (green density, low $\kappa_{\vecX}$) yields a non-specific unknown posterior, i.e., high $\calN_0(\vecX)$.
\textbf{(b)}~Top: the KL summary measures information gain and can invert the risk ordering of two posteriors (A vs.\ B), which is why KL features require the nonlinear calibration of HolUE.
Bottom: MPRisk scores the selected decision with four components $r_{\mathrm{FA}}$, $r_{\mathrm{ID}}$, $r_{\mathrm{FR}}$, and $r_{\mathrm{NS}}=P_0\,\calN_0$; stacked bars show the components at the three marked probes --- for the confidently rejected corrupted probe only $r_{\mathrm{NS}}$ flags the error.
The final score $u_{\lambda}$ combines the components with four nonnegative validation-tuned weights.}
\label{fig:teaser}
\end{figure*}

We propose \emph{MPRisk}, a mixed-prior posterior decision-risk score that avoids the need for nonlinear calibration by scoring the error events of the selected decision directly.
MPRisk reuses the Bayesian posterior over known and unknown classes and combines four components: false-acceptance risk $r_{\mathrm{FA}}$, misidentification risk $r_{\mathrm{ID}}$, false-rejection risk $r_{\mathrm{FR}}$, and a reject non-specificity penalty $r_{\mathrm{NS}}$.
The last component exists only because the unknown part of the class space is modeled as a continuous component representing unknown identities: it asks whether a rejection is supported by a specific unknown-identity hypothesis or merely by diffuse evidence, and thereby detects poor-quality in-gallery probes that are confidently but wrongly rejected.
The ranking score contains no nonlinear fitted model; four nonnegative weights are selected on validation data, and an optional monotone calibration step converts the score into an error-probability estimate for reliability analysis.
Figure~\ref{fig:teaser} summarizes the approach.

We make three contributions:
\begin{itemize}
    \item \textbf{Analysis.} We show that KL-based summaries of the gallery-aware posterior are not generally ordered by the risk of the selected OSR decision, and that linear fusion of the HolUE KL features performs poorly on several benchmarks while nonlinear calibration recovers useful rankings.
    \item \textbf{Method.} We derive MPRisk, which combines posterior terms for false acceptance, misidentification, and false rejection with a closed-form non-specificity penalty for rejected probes based on the continuous unknown component of the mixed prior.
    \item \textbf{Evaluation.} Under matched validation budgets we compare MPRisk with KL-based, post-hoc, and supervised scores on nine image, audio, and text benchmarks, including component ablations, error-type analysis, calibration diagnostics, and runtime measurements.
\end{itemize}

Section~\ref{sec:related} reviews related work; Section~\ref{sec:background} formalizes the problem and the KL--risk mismatch; Section~\ref{sec:method} presents MPRisk; Section~\ref{sec:experiments} reports experiments; Sections~\ref{sec:discussion}--\ref{sec:conclusion} discuss and conclude.

\section{RELATED WORK}\label{sec:related}

\subsection{Open-set recognition and open-world learning}

Open-set recognition was introduced to describe recognition systems that must handle classes absent from training or enrollment~\cite{scheirer2013toward}.
The broader open-world learning literature studies unknown-class rejection, open-set classification, and incremental discovery of new classes~\cite{osr_in_text_review,1811.08581}.
Deep open-set methods such as OpenMax and DOC modify the classifier head or activation statistics to reduce open-space risk~\cite{bendale2016towards,doc}; a further line of work improves the discriminative separation of known and unknown classes with specialized losses and architectures~\cite{osr_in_text_review}.
Most of this literature improves recognition or unknown-detection accuracy, whereas we estimate the reliability of decisions of a \emph{fixed} OSR system without changing the embedding model, gallery, or acceptance rule.

\subsection{Open-set recognition in text}

Several text tasks naturally follow the OSR protocol.
In intent classification a conversational system must reject out-of-scope requests; CLINC150 is the standard benchmark~\cite{clinc150}.
Topic classification datasets such as Yahoo Answers, AGNews, and DBPedia are converted into open-set protocols by treating a subset of topics as known~\cite{yahoo_ostc,20news}.
Authorship attribution and verification are inherently open-set~\cite{stamatatos-2017-authorship,10.1561/1500000005,10.1007/978-3-030-86337-1_15}; PAN protocols provide challenging settings with dynamic galleries~\cite{Kestemont2020OverviewOT,DBLP:journals/corr/abs-2112-05125}.
Our previous work adapted gallery-aware KL-based uncertainty estimation to open-set text classification; here we use both modality groups within the identical Bayesian model.

\subsection{Aleatoric uncertainty and probabilistic embeddings}

Uncertainty in deep learning is commonly decomposed into epistemic and aleatoric components~\cite{kendall2017uncertainties,depeweg2018decomposition,gal2016dropout,lakshminarayanan2017simple}; in our fixed-backbone, fixed-gallery pipeline the dominant source is aleatoric, arising from the relation between a probe and the gallery.
Probabilistic embeddings represent this uncertainty directly: PFE predicts feature uncertainty in Euclidean space~\cite{pfe}, and the SCF model predicts a von Mises--Fisher (vMF) distribution on the unit hypersphere~\cite{scf},
$$
p(\vecZ|\vecX)=C_d(\kappa_{\vecX})\exp(\kappa_{\vecX}\vecM_{\vecX}^\top\vecZ),
$$
where $\vecM_{\vecX}$ is a unit mean direction and $\kappa_{\vecX}$ a concentration~\cite{fisher1993statistical}; large $\kappa_{\vecX}$ corresponds to a high-quality embedding.
MPRisk uses this vMF distribution twice: to average the embedding-conditioned posterior into probe-level class probabilities and to obtain a closed-form non-specificity of the unknown-identity posterior (Section~\ref{sec:ns}).

\subsection{Selective prediction and calibration}

Selective prediction evaluates whether uncertain samples are rejected before confident ones~\cite{1054406,NIPS2017_4a8423d5}.
Classical decision theory prescribes rejecting when the conditional risk of the best action exceeds the rejection cost (Chow's rule); under 0--1 loss the optimal score is the posterior error probability $1-\max_a p(a|\vecX)$, not an entropy- or divergence-based functional.
MPRisk instantiates this principle for the OSR action space, where the loss structure differs across accept and reject decisions.
Calibration, in contrast, concerns whether predicted probabilities match empirical frequencies~\cite{guo2017calibration,naeini2015obtaining,murphy1973proper}; we treat ranking (evaluated by the Prediction Rejection Ratio~\cite{fadeeva2023lmpolygraph}) and probability calibration (evaluated by ECE) as separate goals.

\subsection{Relation to prior work}

We are not aware of prior work that decomposes OSR uncertainty into decision-conditioned posterior risks while retaining a continuous unknown-identity component.
Existing gallery-aware estimators use raw posterior confidence~\cite{my_paper} or information-theoretic summaries with supervised calibration; existing selective-prediction scores operate on logits or sample quality and ignore the accept/reject decision structure.

\section{BACKGROUND AND PROBLEM STATEMENT}\label{sec:background}

\subsection{Open-set recognition protocol}

Let $\mathcal{G}=\{g_1,\ldots,g_K\}$ be a gallery of known classes, each represented by a prototype $\vecM_c\in\sphere$.
At inference the system receives a probe $\vecX$ and applies the fixed decision rule
$$
s(\vecX)=\max_{c\in\{1,\ldots,K\}}\vecM_c^\top\vecZ(\vecX),
\qquad
\vecX\text{ is accepted iff } s(\vecX)\geq\tau,
$$
$$
\hat c(\vecX)=\argmax_{c\in\{1,\ldots,K\}}\vecM_c^\top\vecZ(\vecX)\quad\text{if accepted}.
$$
Uncertainty estimation does not alter this decision; it assigns a score used to decide whether to trust it.
All compared methods therefore share identical recognition performance at zero filtering and differ only in how they order probes for rejection-by-uncertainty.

\subsection{What the uncertainty score should measure}

For selective OSR the ideal score ranks probes by the probability (or cost) of a recognition error \emph{given the decision made}.
If the probe was accepted, the relevant errors are false acceptance and misidentification; if it was rejected, the relevant error is false rejection.
The posterior terms must therefore be conditioned on the action taken: the posterior unknown probability $P_0(\vecX)$ is a false-acceptance risk only for accepted probes, while for rejected probes it is evidence that the rejection was correct.

\subsection{Bayesian mixed-prior posterior}\label{sec:posterior}

We adopt the Bayesian model over embeddings and class labels used by gallery-aware recognition~\cite{my_paper}.
The class variable is mixed:
$$
c\in\{1,\ldots,K\}\cup(K,K+1],
$$
with discrete values for known classes and the continuous interval indexing the continuum of unknown identities.
The prior is
$$
p(c)=\frac{1-\beta}{K}\sum_{i=1}^{K}\delta(c-i)+\beta\,\mathbb{I}\{c\in(K,K+1]\},
$$
where $\delta$ is the Dirac delta function, $\mathbb{I}$ is the indicator function, and $\beta\in(0,1)$ is the prior probability mass assigned to the unknown continuum, i.e., the prior probability that a probe belongs to an identity outside the gallery.
Known-class embedding densities are vMF with shared concentration $\kappa_g$:
$$
p(\vecZ|c=i)=C_d(\kappa_g)\exp(\kappa_g\vecM_i^\top\vecZ),
$$
each unknown identity $c\in(K,K+1]$ has a point embedding $\vecM_c^{\circ}$ distributed uniformly on $\sphere$, so the aggregate unknown embedding density is $p_0(\vecZ)=1/S_{d-1}$, where $S_{d-1}$ is the surface area of $\sphere$, and the marginal is
$$
m(\vecZ)=\frac{1-\beta}{K}\sum_{i=1}^K p(\vecZ|c=i)+\frac{\beta}{S_{d-1}}.
$$
The probe-level posterior integrates over the SCF embedding distribution:
$$
P_i(\vecX)=\int_{\sphere}\frac{\frac{1-\beta}{K}p(\vecZ|c=i)}{m(\vecZ)}\,p(\vecZ|\vecX)\,d\vecZ,
\qquad
P_0(\vecX)=\int_{\sphere}\frac{\beta/S_{d-1}}{m(\vecZ)}\,p(\vecZ|\vecX)\,d\vecZ,
$$
with $P_0(\vecX)+\sum_{i=1}^{K}P_i(\vecX)=1$.

\paragraph{Mean-embedding approximation.}
Following the practical approximation validated for HolUE, we evaluate the embedding-conditioned posterior at the mean embedding $\vecM_{\vecX}$:
$$
P_i(\vecX)\approx\frac{\frac{1-\beta}{K}p(\vecM_{\vecX}|c=i)}{m(\vecM_{\vecX})},
\qquad
P_0(\vecX)\approx\frac{\beta/S_{d-1}}{m(\vecM_{\vecX})}.
$$
This avoids Monte Carlo noise and costs $O(Kd)$ per probe; the concentration $\kappa_{\vecX}$ still enters the method through the non-specificity term of Section~\ref{sec:ns}.
Monte Carlo posterior estimates with up to $32$ samples did not improve PRR in preliminary experiments while multiplying the cost.

\subsection{KL summaries and decision risk}\label{sec:whykl}

HolUE scores each probe by how far the reconstructed posterior $p(c|\vecX)$ has moved away from the prior $p(c)$; we refer to this information-gain quantity, $\KL(p(c|\vecX)\|p(c))$, as the \emph{KL summary} of the posterior.
The summary splits into a gallery term $\mathrm{KL}_1$, accumulated over the known classes, and an unknown-mass term $\mathrm{KL}_2$, accumulated over the continuous unknown component.
HolUE does not use these components directly: it normalizes them on a validation set and fuses them with a supervised nonlinear calibrator trained on validation error labels --- an exponential-transform model in the biometric setting and a small MLP in the text setting.
HolUE is therefore a validation-supervised uncertainty estimator, and throughout this paper it is compared as such.

Why is the nonlinear step needed?
For a uniform prior over $N$ outcomes,
$$
\KL(p\|p_{\mathrm{unif}})=\log N - H(p),
$$
so the KL summary ranks probes by negative posterior entropy, whereas the decision risk of the maximum-posterior action is
$$
R(\vecX)=1-\max_a p(a|\vecX).
$$
For $N\geq 3$, entropy and maximum probability are not monotone functions of each other: a posterior split between two identities --- the misidentification-prone case --- is under-penalized by an entropy-type score relative to a mildly diffuse posterior.
Two further mismatches are specific to OSR.
First, KL aggregates evidence about all actions, whereas risk concerns only the action taken: a rejected probe with a sharp two-way split among gallery classes carries large $\mathrm{KL}_1$ that is irrelevant to the rejection's correctness.
Second, $\mathrm{KL}_2$ grows with the concentration $\kappa_{\vecX}$, so its relation to error probability reverses between accepted and rejected probes.
A fixed monotone transformation of each component cannot in general account for this change of meaning across decision contexts; HolUE therefore learns a flexible nonlinear mapping on validation data.

Table~\ref{tab:kl_rank_inversion} quantifies the raw mismatch on two representative datasets.
Here \emph{Spearman} is the rank correlation between the raw (uncalibrated) KL summary and the per-probe error indicator; the \emph{inversion rate} is the fraction of (erroneous, correct) probe pairs to which the KL summary assigns the wrong order; \emph{KL AUROC} and \emph{KL AUPRC} measure any-error detection quality of the raw KL summary used directly as a score.
The raw summary is nearly uncorrelated with the error ordering on IJB-C and anti-correlated on Yahoo Answers.

\begin{table}[t]
\centering
\footnotesize
\setlength\tabcolsep{4pt}
\caption{Disagreement between the raw (uncalibrated) KL summary and the error-risk ordering. Spearman: rank correlation between the KL summary and the per-probe error indicator. Inv.\ rate: fraction of (erroneous, correct) probe pairs ordered incorrectly by the KL summary. KL AUROC/AUPRC: any-error detection quality of the raw KL summary used directly as an uncertainty score.}
\label{tab:kl_rank_inversion}
\begin{tabular}{lcccc}
\toprule
Dataset & Spearman & Inv. rate & KL AUROC & KL AUPRC \\
\midrule
IJB-C & -0.03 & 0.55 & 0.45 & 0.04 \\
Yahoo Answers & -0.47 & 0.81 & 0.18 & 0.15 \\
\bottomrule
\end{tabular}
\end{table}

\section{MPRISK METHOD}\label{sec:method}

\subsection{Decision-conditioned risk components}\label{sec:components}

Let the fixed OSR decision for probe $\vecX$ be either rejection or acceptance as class $\hat c$.
We call a component \emph{risk-aligned} if, within the decision context in which it is active, it is monotone in the posterior probability of its associated error event.
MPRisk uses three risk-aligned posterior error risks and, in the next subsection, one penalty term.

\paragraph{Acceptance risks.}
If $\vecX$ is accepted as $\hat c$, the probe may be unknown or may belong to a different known class:
$$
r_{\mathrm{FA}}(\vecX)=\mathbb{I}\{\vecX\text{ accepted}\}\,P_0(\vecX),
\qquad
r_{\mathrm{ID}}(\vecX)=\mathbb{I}\{\vecX\text{ accepted}\}\sum_{\substack{j=1\\ j\neq\hat c}}^{K}P_j(\vecX).
$$
For rejected probes both are identically zero.

\paragraph{Rejection risk.}
If $\vecX$ is rejected, an error occurs when the probe belongs to some gallery class:
$$
r_{\mathrm{FR}}(\vecX)=\mathbb{I}\{\vecX\text{ rejected}\}\,\bigl(1-P_0(\vecX)\bigr).
$$

Under equal error costs, $r_{\mathrm{FA}}+r_{\mathrm{ID}}+r_{\mathrm{FR}}$ equals the posterior probability that the taken decision is wrong --- Chow's conditional risk for the OSR action space.

\subsection{Mixed-prior reject non-specificity}\label{sec:ns}

The rejection risk $1-P_0(\vecX)$ is blind to an important failure mode.
For a known probe whose embedding drifted far from the gallery due to corruption, the model assigns $P_0(\vecX)\approx 1$ and the rejection appears maximally confident, although it is a false rejection.
If the unknown class were a single collapsed atom, no posterior quantity could distinguish this case from a genuine unknown.

The mixed prior enables a sharper question: is the unknown explanation specific or diffuse?
Conditioned on the unknown event, the posterior density over unknown identity centers $\vecU\in\sphere$ is
$$
p\bigl(\vecU\,\big|\,\vecX,\,c\in(K,K+1]\bigr)
=\frac{\dfrac{\beta}{S_{d-1}}\,\dfrac{p(\vecU|\vecX)}{m(\vecU)}}{P_0(\vecX)}.
$$
A high-quality unknown probe induces a concentrated posterior over latent unknown identity centers, whereas a corrupted known probe induces a diffuse one: its large $P_0$ then reflects poor embedding quality rather than evidence for a coherent unknown identity.
We quantify concentration by the normalized collision probability
$$
\mathcal{C}_0(\vecX)=S_{d-1}\int_{\sphere}p\bigl(\vecU|\vecX,c\in(K,K+1]\bigr)^2\,d\vecU,
\qquad
\calN_0(\vecX)=\frac{1}{\mathcal{C}_0(\vecX)}\in(0,1],
$$
so that $\calN_0\to 1$ for a fully non-specific unknown posterior and $\calN_0\to 0$ for a concentrated one.
The \emph{non-specificity penalty} is
$$
r_{\mathrm{NS}}(\vecX)=\mathbb{I}\{\vecX\text{ rejected}\}\;P_0(\vecX)\,\calN_0(\vecX):
$$
a rejection is flagged when it places large mass on the unknown component without committing to any specific unknown identity.
Unlike the three error risks, $r_{\mathrm{NS}}$ is not itself a posterior error probability; we refer to the four terms collectively as \emph{components}.

\paragraph{Closed form.}
Approximating the unknown-identity posterior by the vMF embedding distribution predicted by SCF, the collision integral of a vMF density is analytic:
$$
\int_{\sphere}p(\vecZ|\vecX)^2 d\vecZ=\frac{C_d(\kappa_{\vecX})^2}{C_d(2\kappa_{\vecX})},
\qquad\Longrightarrow\qquad
\calN_0(\vecX)\approx\frac{C_d(2\kappa_{\vecX})}{S_{d-1}\,C_d(\kappa_{\vecX})^2}.
$$
Low-quality samples have small $\kappa_{\vecX}$ and hence large $\calN_0$.
Note that $\calN_0$ alone is a monotone transform of the SCF quality score; the value added by $r_{\mathrm{NS}}$ lies in its decision gating and weighting by $P_0$, which the ablations of Section~\ref{sec:mixedprior} isolate.

\subsection{Final score, weight tuning, and probability calibration}\label{sec:tuning}

The MPRisk score is a nonnegative combination of the four components:
$$
u_{\lambda}(\vecX)=
\lambda_{\mathrm{FA}}\,r_{\mathrm{FA}}(\vecX)
+\lambda_{\mathrm{ID}}\,r_{\mathrm{ID}}(\vecX)
+\lambda_{\mathrm{FR}}\,r_{\mathrm{FR}}(\vecX)
+\lambda_{\mathrm{NS}}\,r_{\mathrm{NS}}(\vecX).
$$
The equal-cost version, $\lambda\equiv 1$, is reported as \emph{MPRisk raw}.
The tuned version selects
$$
\lambda^{\star}=\argmax_{\lambda\in\R_{+}^{4}}\ \mathrm{PRR}^{F_1}_{\mathrm{val}}(u_\lambda)
$$
on a disjoint validation set by a randomized search over log-uniform weight candidates (the score is scale-invariant in $\lambda$).
The weights can be interpreted as relative costs assigned to the components; different datasets and FPIR operating points assign different effective importance to the error types (Section~\ref{sec:errortypes}), so cost selection is part of the method.
The four-parameter score has substantially lower capacity than the supervised baselines considered in Section~\ref{sec:fair}.
Note that weight tuning uses validation error labels through the PRR objective; what MPRisk avoids is a nonlinear fitted model in the ranking pipeline.

An optional scalar monotone calibrator (fitted on validation error indicators) maps $u_{\lambda^\star}$ to a calibrated error probability; this variant, \emph{MPRisk cal}, preserves the ranking up to ties and is used for reliability analysis only.

\subsection{Computational cost}

All components are functions of quantities already computed for the recognition decision plus the analytic $\calN_0$; the per-probe overhead is $O(Kd)$ for the posterior and $O(1)$ for the risk formula.
MPRisk raw requires no tuning at all; weight tuning and probability calibration are offline, one-off procedures on the validation set.
Measured runtimes are reported in Section~\ref{sec:runtime}.

\section{EXPERIMENTS}\label{sec:experiments}

\subsection{Datasets, protocols, and baselines}

We evaluate on four image and audio identification benchmarks --- IJB-C~\cite{ijbc}, IJB-B~\cite{ijbb}, Whale~\cite{cheeseman2022happywhale}, and VoxBlink (VB-Eval-L-5)~\cite{voxblink} --- and five text benchmarks --- Yahoo Answers, AGNews, DBPedia~\cite{yahoo_ostc}, CLINC150~\cite{clinc150}, and PAN-20-AV~\cite{Kestemont2020OverviewOT}.
Backbones, galleries, thresholds, and OSR protocols follow our previous work exactly: ArcFace/SCF backbones for image and audio, the Whale OSR protocol built on HappyWhale, and frozen-BERT SCF heads with the gallery constructions of our previous work for text.
Recognition decisions are fixed throughout; methods are compared purely as uncertainty estimators.
Validation splits are disjoint from test in identities/authors/topics and are used for all weight tuning, supervised fitting, and calibration; every validation-consuming method receives the same validation data.

Baselines include SCF (sample quality), AccScr (distance to the rejection threshold), MSP and Margin (post-hoc confidence over gallery-augmented logits with calibrated temperature), GalUE (gallery ambiguity), and HolUE in its original validation-calibrated form (Section~\ref{sec:whykl})~\cite{my_paper}.
The prior mass is $\beta=0.5$ everywhere; robustness of the shared posterior to $\beta$ was established previously and carries over to MPRisk.

\subsection{Metrics}

Our primary metric is the Prediction Rejection Ratio (PRR)~\cite{fadeeva2023lmpolygraph}, computed from the $F_1$ score as probes are removed in decreasing order of uncertainty:
$$
\mathrm{PRR}=\frac{\mathrm{AUC}_{\mathrm{unc}}-\mathrm{AUC}_{\mathrm{random}}}{\mathrm{AUC}_{\mathrm{oracle}}-\mathrm{AUC}_{\mathrm{random}}},
$$
so that $1$ corresponds to oracle filtering, $0$ to random filtering, and negative values to a score worse than random.
We complement PRR with per-error-type detection AUROC (any error, FA, FR, ID), paired bootstrap confidence intervals for PRR differences (resampling over probes), expected calibration error (ECE) with reliability diagrams for the calibrated variants, and wall-clock per-probe runtime.

\subsection{Compared estimators under a fixed validation budget}\label{sec:tunedbaselines}

Both HolUE and tuned MPRisk consume the same validation set, so the head-to-head comparison below is tuned-versus-tuned.
To isolate what the feature representation contributes beyond the shared validation access, we additionally evaluate a family of estimators built on the same posterior and validation data:
\begin{itemize}
    \item \textbf{Linear KL fusion}: the KL evidence restricted to a linear rule.
    The features $-\mathrm{KL}_1$, $-\mathrm{KL}_2$, and $-(\mathrm{KL}_1+\mathrm{KL}_2)$ are sign-expanded (each feature also enters negated, so arbitrary signed combinations are reachable), standardized on validation, and combined with weights selected by the same randomized $\mathrm{PRR}^{F_1}_{\mathrm{val}}$ search used for $\lambda^{\star}$.
    Comparing this row with HolUE isolates the contribution of the nonlinear calibrator; comparing it with tuned MPRisk isolates the contribution of the risk features under identical linear tuning.
    \item \textbf{Tuned simple}: the same linear protocol applied to the post-hoc scores $\{$SCF, AccScr, MSP, Margin$\}$.
    \item \textbf{Rejected-SCF}: the SCF quality score gated by the rejection decision (accepted probes receive a constant low value), isolating the gating idea from the $P_0$ weighting in $r_{\mathrm{NS}}$.
    \item \textbf{Supervised logistic / Supervised MLP}: error detectors trained on validation with binary any-error labels over a $14$-dimensional feature vector combining the four post-hoc scores with all posterior features ($-\mathrm{KL}_1$, $-\mathrm{KL}_2$, their sum, $P_0$, $\calN_0$, the four MPRisk components, and the raw MPRisk sum).
    The logistic model uses balanced class weights; the MLP is a single-hidden-layer network (16 units, weight decay, early stopping).
    These are stronger supervised baselines with a larger feature set; note they differ from HolUE's calibrator, which operates on only two normalized features with a task-specific objective.
    \item \textbf{Hybrid KL+MPRisk}: a tuned linear combination of the Linear-KL-fusion score and the tuned MPRisk score, testing whether information gain and decision risk are complementary.
\end{itemize}

\subsection{Main results: image, Whale, and audio}\label{sec:mainbio}

\begin{figure*}[t]
\centering
\includegraphics[width=0.32\linewidth]{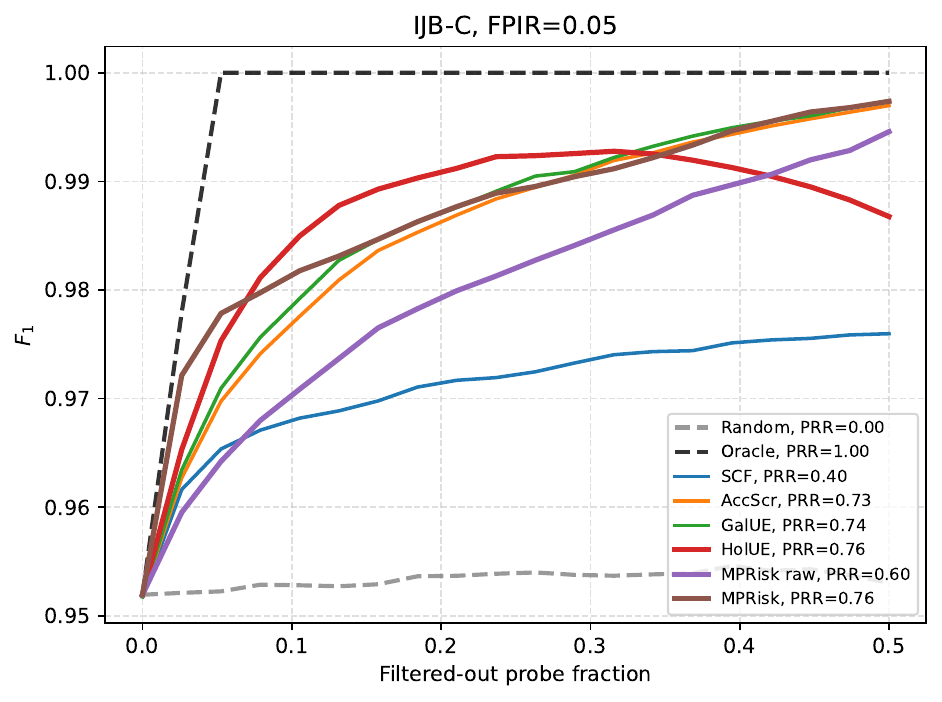}\hfill
\includegraphics[width=0.32\linewidth]{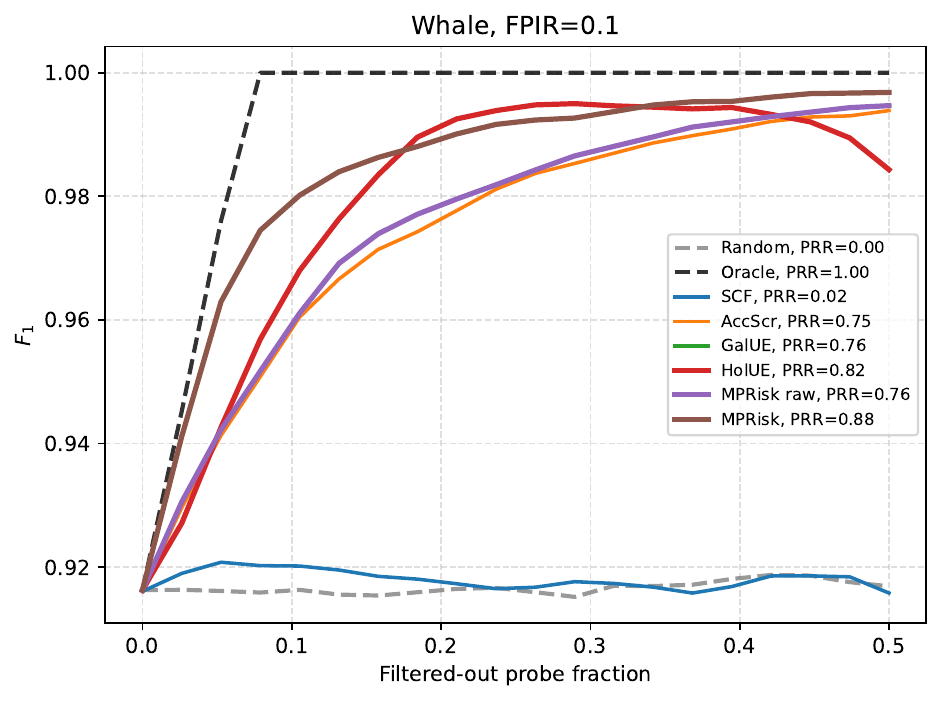}\hfill
\includegraphics[width=0.32\linewidth]{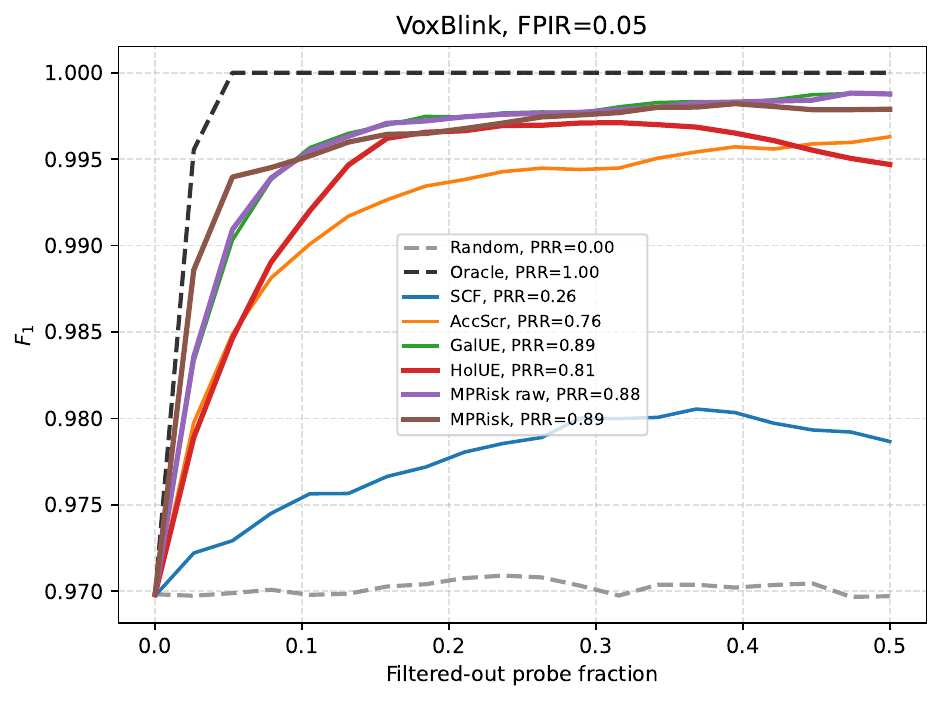}
\caption{Risk-controlled filtering curves on IJB-C at FPIR $0.05$ (left), Whale at FPIR $0.1$ (middle), and VoxBlink at FPIR $0.05$ (right).
Each panel shows $F_1$, FPIR, and FNIR versus the fraction of filtered probes; probes are removed in decreasing order of uncertainty; PRR values are given in the legends, with Random and Oracle curves as references.
MPRisk tracks the Oracle most closely on $F_1$; unlike sample-quality scores (SCF) it reduces FPIR rapidly, and unlike gallery-only scores (AccScr, GalUE) it also reduces FNIR by filtering confident non-specific rejections early.
Best viewed zoomed in.}
\label{fig:rejection_bio}
\end{figure*}

Table~\ref{tab:mprisk_main_prr_bio} reports PRR on IJB-C, IJB-B, Whale, and VoxBlink at three FPIR operating points each; Figure~\ref{fig:rejection_bio} shows the corresponding rejection curves.

\begin{table}[t]
\centering
\footnotesize
\setlength\tabcolsep{2pt}
\caption{Prediction Rejection Ratios (PRR, $\uparrow$) for $F_1$ filtering on image, Whale, and audio open-set recognition benchmarks. HolUE uses its validation-trained calibration; MPRisk uses validation-tuned component weights; both receive identical validation data. Best results in bold, second-best underlined.}
\label{tab:mprisk_main_prr_bio}
\begin{tabular}{lcccccccccccc}
\toprule
Method & \multicolumn{3}{c}{IJB-C} & \multicolumn{3}{c}{IJB-B} & \multicolumn{3}{c}{Whale} & \multicolumn{3}{c}{VB-Eval-L-5} \\
 & $0.05$ & $0.1$ & $0.2$ & $0.05$ & $0.1$ & $0.2$ & $0.05$ & $0.1$ & $0.2$ & $0.01$ & $0.05$ & $0.1$ \\
\midrule
SCF & 0.40 & 0.31 & 0.23 & 0.29 & 0.25 & 0.22 & 0.16 & 0.02 & -0.06 & 0.55 & 0.26 & 0.12 \\
AccScr & 0.73 & 0.72 & 0.66 & 0.65 & 0.68 & 0.62 & 0.77 & 0.75 & 0.66 & 0.66 & 0.76 & 0.72 \\
MSP & \underline{0.74} & 0.75 & 0.70 & \underline{0.66} & 0.70 & 0.65 & 0.77 & 0.77 & 0.70 & 0.38 & \underline{0.88} & 0.86 \\
Margin & \underline{0.74} & 0.75 & 0.70 & \underline{0.66} & 0.70 & \underline{0.66} & 0.77 & 0.77 & 0.70 & 0.68 & \underline{0.88} & 0.86 \\
GalUE & \underline{0.74} & 0.74 & 0.67 & \underline{0.66} & 0.69 & 0.60 & 0.78 & 0.76 & 0.70 & 0.69 & \textbf{0.89} & 0.87 \\
HolUE & \textbf{0.76} & \underline{0.81} & \underline{0.73} & 0.54 & \underline{0.71} & 0.63 & \underline{0.79} & \underline{0.82} & \underline{0.83} & \underline{0.74} & 0.81 & \underline{0.89} \\
MPRisk raw (ours) & 0.60 & 0.44 & 0.31 & 0.54 & 0.38 & 0.29 & 0.78 & 0.76 & 0.70 & 0.65 & \underline{0.88} & 0.87 \\
MPRisk (ours) & \textbf{0.76} & \textbf{0.83} & \textbf{0.91} & \textbf{0.67} & \textbf{0.76} & \textbf{0.87} & \textbf{0.81} & \textbf{0.88} & \textbf{0.94} & \textbf{0.75} & \textbf{0.89} & \textbf{0.94} \\
\bottomrule
\end{tabular}
\end{table}

MPRisk is best or tied-best at all 12 reported operating points.
The gains grow with FPIR: at FPIR $0.2$ MPRisk reaches $0.91$ on IJB-C ($+0.18$ over HolUE), $0.87$ on IJB-B, and $0.94$ on Whale ($+0.11$ over HolUE) --- higher FPIR admits more unknown probes, so the acceptance-conditioned risks carry more of the error mass.
MPRisk raw is substantially weaker than the tuned version (e.g., $0.44$ vs.\ $0.83$ on IJB-C at FPIR $0.1$): the equal-cost sum is dominated by the numerous small rejection risks, so cost selection is integral to the method.

\subsection{Main results: text}\label{sec:maintext}

\begin{figure*}[t]
\centering
\includegraphics[width=0.33\linewidth]{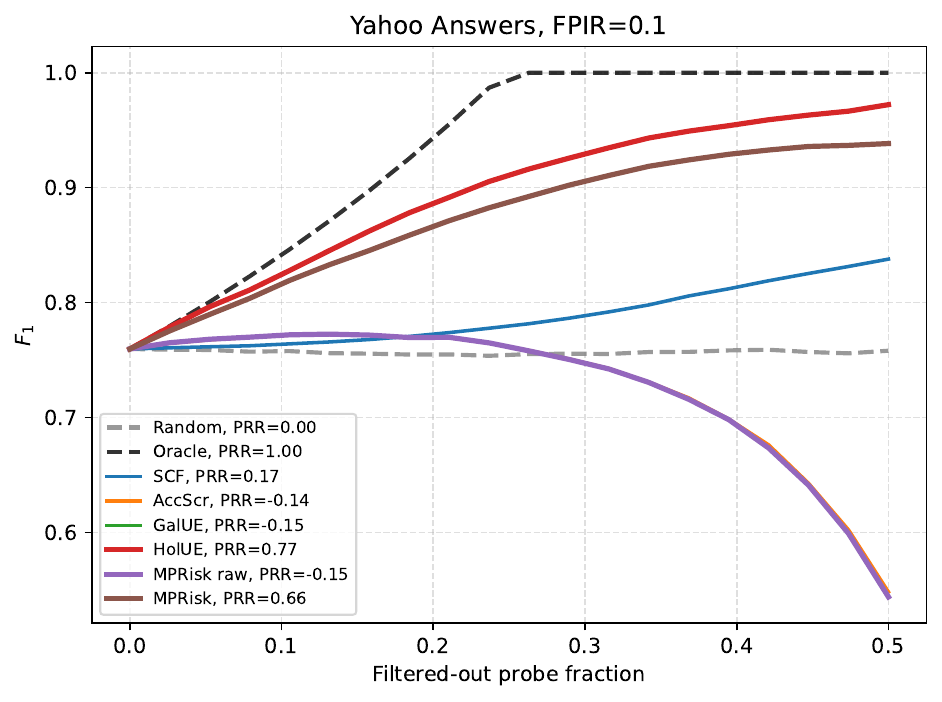}\hfill
\includegraphics[width=0.33\linewidth]{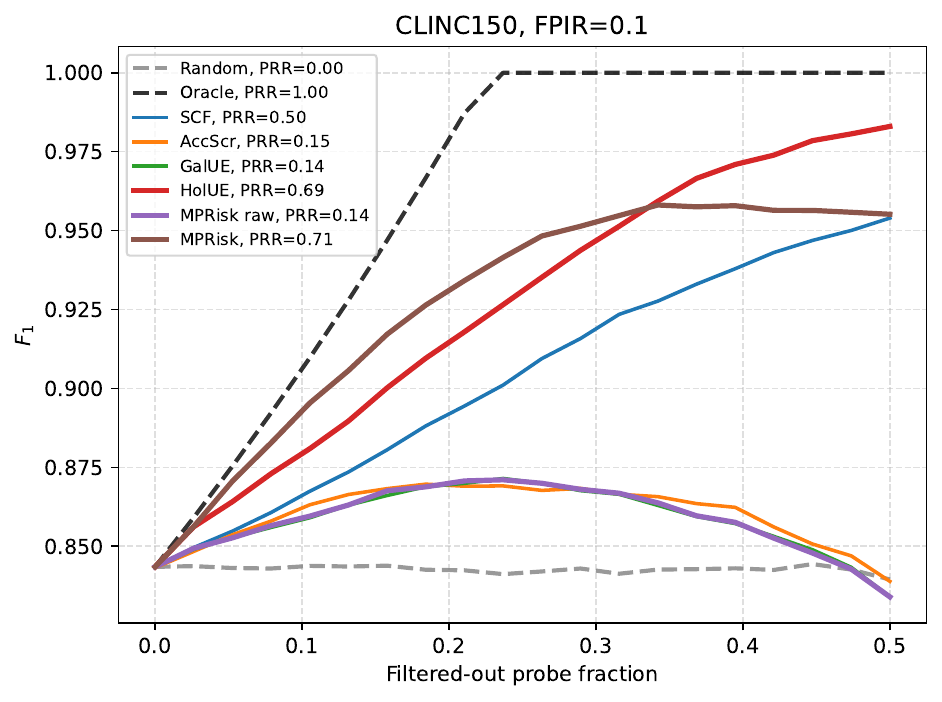}\hfill
\includegraphics[width=0.33\linewidth]{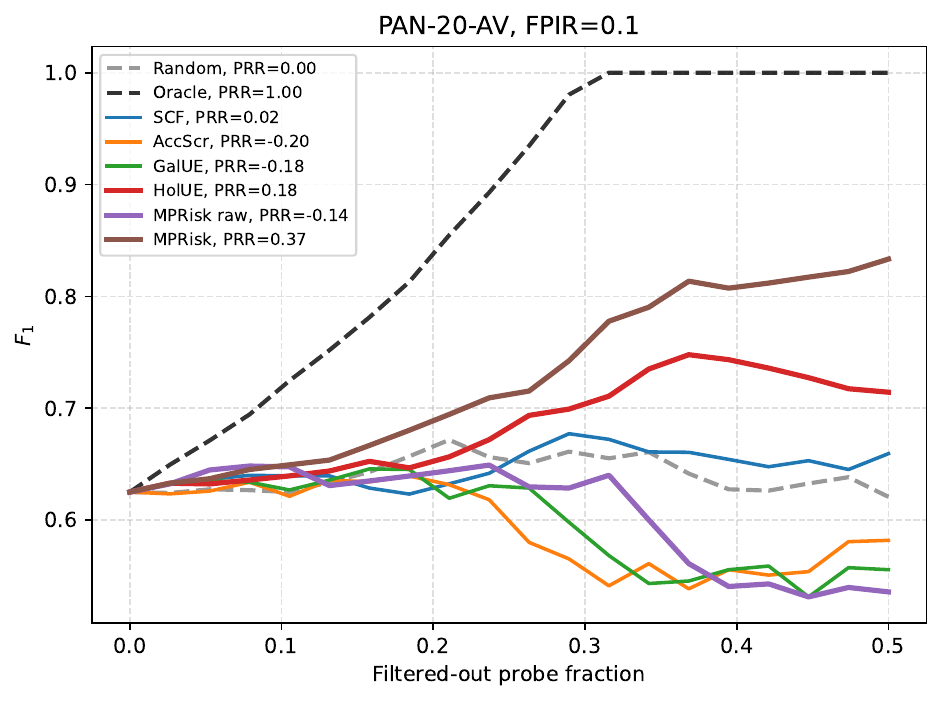}
\caption{Risk-controlled filtering curves on the text benchmarks at FPIR $0.1$: Yahoo Answers, CLINC150, and PAN-20-AV (left to right).
Each panel shows $F_1$, FPIR, and FNIR versus filter-out rate, with PRR in the legends and Random/Oracle references.
On CLINC150 and PAN-20-AV, MPRisk dominates the $F_1$ curve and lowers FPIR fastest; on Yahoo Answers, HolUE's curve is higher because the error mass consists almost entirely of false rejections (cf.\ Table~\ref{tab:error_type_detection_text}).
Best viewed zoomed in.}
\label{fig:rejection_text}
\end{figure*}

Table~\ref{tab:mprisk_main_prr_text} reports PRR on the five text benchmarks at five operating points each; Figure~\ref{fig:rejection_text} shows rejection curves.

\begin{table}[t]
\centering
\footnotesize
\setlength\tabcolsep{2pt}
\caption{Prediction Rejection Ratios (PRR, $\uparrow$) for $F_1$ filtering on open-set text classification benchmarks. HolUE uses its validation-trained MLP calibration; MPRisk uses validation-tuned component weights. Best results in bold, second-best underlined.}
\label{tab:mprisk_main_prr_text}
\begin{adjustbox}{width=\textwidth}
\begin{tabular}{lccccccccccccccccccccccccc}
\toprule
Method & \multicolumn{5}{c}{Yahoo Answers} & \multicolumn{5}{c}{AGNews} & \multicolumn{5}{c}{DBPedia} & \multicolumn{5}{c}{CLINC150} & \multicolumn{5}{c}{PAN-20-AV} \\
 & $0.1$ & $0.2$ & $0.3$ & $0.4$ & $0.5$ & $0.1$ & $0.2$ & $0.3$ & $0.4$ & $0.5$ & $0.1$ & $0.2$ & $0.3$ & $0.4$ & $0.5$ & $0.1$ & $0.2$ & $0.3$ & $0.4$ & $0.5$ & $0.1$ & $0.2$ & $0.3$ & $0.4$ & $0.5$ \\
\midrule
SCF & 0.17 & 0.28 & 0.40 & \underline{0.49} & 0.55 & -0.07 & -0.07 & -0.14 & -0.14 & -0.11 & 0.18 & 0.30 & 0.41 & 0.48 & 0.56 & 0.50 & 0.49 & 0.47 & 0.46 & 0.48 & 0.02 & 0.03 & 0.05 & 0.20 & 0.23 \\
AccScr & -0.14 & 0.28 & 0.43 & 0.48 & 0.49 & -0.02 & 0.39 & \underline{0.48} & 0.51 & 0.50 & 0.39 & \underline{0.78} & \underline{0.83} & \underline{0.84} & 0.79 & 0.15 & 0.40 & 0.49 & \underline{0.56} & \textbf{0.59} & -0.20 & 0.31 & 0.44 & 0.37 & 0.43 \\
MSP & 0.24 & 0.35 & 0.41 & 0.43 & 0.45 & 0.06 & \underline{0.43} & \underline{0.48} & 0.51 & 0.19 & 0.39 & \underline{0.78} & \underline{0.83} & \underline{0.84} & 0.79 & 0.68 & \underline{0.62} & \textbf{0.59} & 0.53 & \underline{0.55} & 0.17 & \underline{0.39} & \underline{0.50} & \underline{0.50} & 0.59 \\
Margin & -0.14 & 0.28 & 0.43 & 0.48 & 0.49 & -0.02 & 0.39 & \underline{0.48} & 0.51 & 0.50 & 0.39 & \underline{0.78} & \underline{0.83} & \underline{0.84} & 0.79 & 0.14 & 0.36 & 0.40 & 0.39 & 0.38 & -0.23 & 0.31 & 0.46 & 0.42 & 0.51 \\
GalUE & -0.15 & 0.27 & 0.43 & 0.48 & 0.49 & -0.03 & 0.38 & 0.47 & 0.51 & 0.51 & \textbf{0.77} & \underline{0.78} & \underline{0.83} & \underline{0.84} & 0.79 & 0.14 & 0.35 & 0.39 & 0.38 & 0.37 & -0.18 & 0.37 & 0.46 & 0.45 & 0.53 \\
HolUE & \textbf{0.77} & \textbf{0.71} & \textbf{0.69} & \textbf{0.71} & \textbf{0.74} & \underline{0.47} & \textbf{0.45} & \textbf{0.52} & \textbf{0.62} & \underline{0.71} & 0.52 & 0.72 & 0.81 & \textbf{0.94} & \textbf{0.96} & \underline{0.69} & \underline{0.62} & \underline{0.57} & \underline{0.56} & \textbf{0.59} & \underline{0.18} & 0.35 & 0.40 & 0.49 & 0.43 \\
MPRisk raw (ours) & -0.15 & 0.27 & 0.43 & 0.48 & 0.49 & 0.02 & 0.38 & 0.47 & 0.51 & 0.51 & 0.65 & \underline{0.78} & \underline{0.83} & 0.83 & 0.79 & 0.14 & 0.36 & 0.40 & 0.40 & 0.41 & -0.14 & \textbf{0.43} & \textbf{0.51} & \textbf{0.53} & \underline{0.60} \\
MPRisk (ours) & \underline{0.66} & \underline{0.52} & \underline{0.46} & 0.40 & \underline{0.64} & \textbf{0.54} & 0.39 & \underline{0.48} & \underline{0.58} & \textbf{0.76} & \underline{0.72} & \textbf{0.85} & \textbf{0.92} & \textbf{0.94} & \underline{0.95} & \textbf{0.71} & \textbf{0.65} & \textbf{0.59} & \textbf{0.57} & \textbf{0.59} & \textbf{0.37} & \textbf{0.43} & 0.48 & 0.44 & \textbf{0.62} \\
\bottomrule
\end{tabular}
\end{adjustbox}
\end{table}

MPRisk performs best on DBPedia (e.g., $0.92$ vs.\ $0.81$ for HolUE at FPIR $0.3$), CLINC150 (best at all five points), and PAN-20-AV at most points (e.g., $0.37$ vs.\ $0.18$ at FPIR $0.1$), and takes the extreme operating points on AGNews.
HolUE is consistently stronger on Yahoo Answers.
The error-type analysis in Section~\ref{sec:errortypes} explains the exception: Yahoo errors are almost exclusively false rejections, a regime in which the quality-driven $\mathrm{KL}_2$ feature under HolUE's calibration is a near-ideal specialist.
This dataset-level complementarity motivates the hybrid estimator studied in Section~\ref{sec:fair}.

\subsection{Feature-space study}\label{sec:fair}

Tables~\ref{tab:fair_tuning_bio} and~\ref{tab:fair_tuning_text} evaluate the estimator family of Section~\ref{sec:tunedbaselines}: all rows share the validation split, the tuning objective $\mathrm{PRR}^{F_1}_{\mathrm{val}}$, and (for the linear rows) the same randomized search budget.

\begin{table}[t]
\centering
\scriptsize
\setlength\tabcolsep{2pt}
\caption{Feature-space study on image, Whale, and audio benchmarks (PRR, $\uparrow$). All rows use the same validation split and, where applicable, the same tuning objective and search budget. Linear KL fusion applies MPRisk's linear tuning protocol to the KL components (with sign expansion); it should be contrasted with HolUE in Table~\ref{tab:mprisk_main_prr_bio}, which fuses the same KL components with a supervised nonlinear calibrator. MPRisk rows correspond to an independent re-run of the $\lambda$ search and may differ from Table~\ref{tab:mprisk_main_prr_bio} by $\pm 0.01$.}
\label{tab:fair_tuning_bio}
\begin{adjustbox}{width=\textwidth}
\begin{tabular}{lcccccccccccc}
\toprule
Method & \multicolumn{3}{c}{IJB-C} & \multicolumn{3}{c}{IJB-B} & \multicolumn{3}{c}{Whale} & \multicolumn{3}{c}{VB-Eval-L-5} \\
 & $0.05$ & $0.1$ & $0.2$ & $0.05$ & $0.1$ & $0.2$ & $0.05$ & $0.1$ & $0.2$ & $0.01$ & $0.05$ & $0.1$ \\
\midrule
MPRisk raw & 0.6 & 0.44 & 0.31 & 0.54 & 0.38 & 0.29 & 0.78 & 0.76 & 0.7 & 0.65 & \underline{0.88} & 0.87 \\
MPRisk tuned no $r_{NS}$ & \underline{0.76} & \underline{0.84} & 0.9 & \underline{0.66} & 0.77 & 0.87 & 0.82 & 0.89 & \underline{0.94} & 0.71 & 0.82 & \textbf{0.95} \\
MPRisk & \underline{0.76} & \underline{0.84} & \underline{0.91} & \underline{0.66} & 0.77 & \underline{0.88} & 0.81 & 0.89 & \underline{0.94} & 0.75 & \textbf{0.89} & \textbf{0.95} \\
Linear KL fusion & -0.67 & -0.5 & 0.17 & -1.31 & -0.18 & 0.49 & 0.61 & 0.6 & 0.61 & 0.58 & 0.31 & 0.43 \\
Tuned simple & 0.7 & 0.82 & 0.84 & -2.89 & -2.34 & -1.88 & 0.8 & 0.81 & 0.76 & 0.77 & \textbf{0.89} & 0.87 \\
Rejected-SCF & 0.28 & 0.15 & 0.08 & 0.29 & 0.16 & 0.08 & 0.27 & 0.11 & 0.04 & 0.68 & 0.07 & 0 \\
Supervised logistic & 0.7 & 0.83 & \underline{0.91} & \underline{0.66} & \underline{0.78} & 0.87 & \textbf{0.86} & \textbf{0.91} & \textbf{0.95} & \textbf{0.88} & \textbf{0.89} & 0.91 \\
Supervised MLP & -0.7 & 0.23 & 0.67 & 0.56 & 0.67 & 0.82 & -0.38 & 0.46 & 0.88 & 0.34 & 0.01 & -0.03 \\
Hybrid KL+MPRisk & \textbf{0.79} & \textbf{0.87} & \textbf{0.93} & \textbf{0.68} & \textbf{0.8} & \textbf{0.89} & \underline{0.84} & \underline{0.9} & \textbf{0.95} & \underline{0.83} & \textbf{0.89} & \underline{0.93} \\
\bottomrule
\end{tabular}
\end{adjustbox}
\end{table}

\begin{table}[t]
\centering
\scriptsize
\setlength\tabcolsep{2pt}
\caption{Feature-space study on open-set text benchmarks (PRR, $\uparrow$). Conventions as in Table~\ref{tab:fair_tuning_bio}; the HolUE reference (KL features with supervised MLP calibration) is given in Table~\ref{tab:mprisk_main_prr_text}.}
\label{tab:fair_tuning_text}
\begin{adjustbox}{width=\textwidth}
\begin{tabular}{lccccccccccccccc}
\toprule
Method & \multicolumn{3}{c}{Yahoo Answers} & \multicolumn{3}{c}{AGNews} & \multicolumn{3}{c}{DBPedia} & \multicolumn{3}{c}{CLINC150} & \multicolumn{3}{c}{PAN-20-AV} \\
 & $0.1$ & $0.3$ & $0.5$ & $0.1$ & $0.3$ & $0.5$ & $0.1$ & $0.3$ & $0.5$ & $0.1$ & $0.3$ & $0.5$ & $0.1$ & $0.3$ & $0.5$ \\
\midrule
MPRisk raw & -0.15 & 0.43 & 0.49 & 0.02 & 0.47 & 0.51 & 0.65 & 0.83 & 0.79 & 0.14 & 0.4 & 0.41 & -0.14 & \underline{0.51} & 0.6 \\
MPRisk tuned no $r_{NS}$ & 0.66 & 0.51 & 0.64 & \textbf{0.54} & 0.48 & \underline{0.76} & \underline{0.72} & \underline{0.92} & \underline{0.95} & 0.69 & \underline{0.6} & 0.59 & \underline{0.33} & 0.49 & \underline{0.64} \\
MPRisk & 0.66 & 0.46 & 0.64 & \underline{0.53} & 0.48 & \underline{0.76} & \underline{0.72} & \underline{0.92} & \underline{0.95} & 0.71 & 0.59 & 0.59 & 0.31 & 0.47 & 0.61 \\
Linear KL fusion & 0.71 & 0.36 & 0.51 & 0.5 & 0.43 & 0.41 & 0.44 & 0.42 & 0.53 & 0.64 & 0.55 & 0.55 & 0.26 & 0.42 & 0.5 \\
Tuned simple & 0.33 & 0.5 & 0.72 & 0.07 & 0.16 & 0.27 & 0.17 & 0.22 & 0.43 & 0.69 & \underline{0.6} & \textbf{0.63} & 0.31 & 0.46 & 0.47 \\
Rejected-SCF & 0.18 & 0.18 & -0.07 & 0.22 & 0.01 & -0.22 & 0.05 & -0.08 & -0.21 & 0.59 & 0.27 & 0 & 0.25 & 0.22 & 0.28 \\
Supervised logistic & \underline{0.73} & \underline{0.68} & \underline{0.75} & 0.32 & \textbf{0.62} & 0.75 & \textbf{0.89} & \textbf{0.93} & \textbf{0.97} & \textbf{0.76} & \textbf{0.62} & \underline{0.61} & \underline{0.33} & 0.45 & 0.61 \\
Supervised MLP & \textbf{0.88} & \textbf{0.86} & \textbf{0.86} & \underline{0.53} & -0.37 & 0.72 & \textbf{0.89} & 0.8 & \underline{0.95} & \textbf{0.76} & 0.58 & -0.18 & 0.19 & -0.33 & -0.22 \\
Hybrid KL+MPRisk & \underline{0.73} & 0.49 & 0.63 & \textbf{0.54} & \underline{0.57} & \textbf{0.79} & 0.71 & \underline{0.92} & \underline{0.95} & \underline{0.74} & 0.59 & 0.59 & \textbf{0.43} & \textbf{0.61} & \textbf{0.67} \\
\bottomrule
\end{tabular}
\end{adjustbox}
\end{table}

Four conclusions follow.

\textbf{(i) Linear fusion is effective for MPRisk features but not for the KL features.}
Linear KL fusion produces negative PRR on IJB-C ($-0.67$/$-0.50$) and IJB-B ($-1.31$) and trails elsewhere, whereas the same KL evidence fused by HolUE's nonlinear calibrator reaches $0.76$/$0.81$ on IJB-C (Table~\ref{tab:mprisk_main_prr_bio}).
This is the analysis of Section~\ref{sec:whykl} made quantitative.
MPRisk reaches $0.76$/$0.84$/$0.91$ on IJB-C with four nonnegative weights: the risk decomposition yields features that combine effectively under a linear score.

\textbf{(ii) Linear fusion of heterogeneous post-hoc scores generalizes poorly.}
Tuned simple is competitive on IJB-C and VoxBlink but substantially negative on IJB-B ($-2.89$) and weak on DBPedia; scores on incompatible scales transfer poorly from validation to test even after standardization, whereas the MPRisk components share a common posterior-probability scale.

\textbf{(iii) MPRisk is competitive with supervised detectors at much lower capacity.}
Supervised logistic --- trained on error labels over the full $14$-dimensional feature set --- exceeds tuned MPRisk only modestly (e.g., Whale $0.86$ vs.\ $0.81$ at FPIR $0.05$; DBPedia $0.89$ vs.\ $0.72$ at FPIR $0.1$) and loses at some points (VoxBlink at FPIR $0.1$).
The supervised MLP varies substantially across datasets, from $0.88$ on Yahoo to $-0.70$ on IJB-C and $-0.38$ on Whale; this does not contradict the stability of HolUE's two-input, task-objective MLP --- on moderate validation sets, performance tracks how tightly the hypothesis class matches the task rather than raw capacity.

\textbf{(iv) Decision risk and information gain are complementary.}
Hybrid KL+MPRisk, a tuned combination of two scalars, obtains the highest PRR at $9$ of $12$ image/audio operating points and on AGNews and PAN-20-AV (e.g., $0.43$ at PAN FPIR $0.1$, $+0.25$ over HolUE); on Yahoo it recovers most of HolUE's advantage ($0.73$ vs.\ $0.77$).

\subsection{Component ablation}\label{sec:ablation}

Tables~\ref{tab:full_component_ablation_bio} and~\ref{tab:full_component_ablation_text} decompose MPRisk into its parts.

\begin{table}[t]
\centering
\scriptsize
\setlength\tabcolsep{2pt}
\caption{Full MPRisk component ablation on image, Whale, and audio benchmarks. PRR for $F_1$ filtering is reported.}
\label{tab:full_component_ablation_bio}
\begin{adjustbox}{width=\textwidth}
\begin{tabular}{lcccccccccccc}
\toprule
Variant & \multicolumn{3}{c}{IJB-C} & \multicolumn{3}{c}{IJB-B} & \multicolumn{3}{c}{Whale} & \multicolumn{3}{c}{VB-Eval-L-5} \\
 & $0.05$ & $0.1$ & $0.2$ & $0.05$ & $0.1$ & $0.2$ & $0.05$ & $0.1$ & $0.2$ & $0.01$ & $0.05$ & $0.1$ \\
\midrule
$r_{FA}$ & -0.92 & 0.34 & 0.76 & -1.05 & 0.22 & 0.72 & -3.17 & -0.63 & \underline{0.84} & -3.54 & 0.53 & 0.85 \\
$r_{ID}$ & -0.91 & 0.35 & 0.77 & -1.06 & 0.22 & 0.73 & -3.21 & -0.66 & 0.8 & -0.65 & 0.56 & 0.87 \\
$r_{FR}$ & 0.22 & 0.11 & 0.05 & 0.25 & 0.13 & 0.06 & 0.26 & 0.1 & 0.03 & \underline{0.8} & 0.08 & 0.01 \\
$r_{NS}$ & 0.28 & 0.15 & 0.06 & 0.29 & 0.16 & 0.08 & 0.27 & 0.11 & 0.04 & 0.67 & 0.07 & 0 \\
$r_{FA}+r_{ID}$ & -0.91 & 0.35 & 0.77 & -1.05 & 0.22 & 0.73 & -3.18 & -0.63 & 0.83 & -3.52 & 0.59 & \underline{0.88} \\
$r_{FR}+r_{NS}$ & 0.22 & 0.11 & 0.05 & 0.25 & 0.13 & 0.06 & 0.26 & 0.1 & 0.03 & \textbf{0.81} & 0.08 & 0.01 \\
Ordinary no $r_{NS}$ & \underline{0.6} & \underline{0.44} & 0.31 & \underline{0.54} & \underline{0.38} & 0.29 & 0.78 & \underline{0.76} & 0.7 & 0.65 & \underline{0.88} & 0.87 \\
Full raw & \underline{0.6} & \underline{0.44} & 0.31 & \underline{0.54} & \underline{0.38} & 0.29 & 0.78 & \underline{0.76} & 0.7 & 0.65 & \underline{0.88} & 0.87 \\
Tuned no $r_{NS}$ & \textbf{0.76} & \textbf{0.84} & \underline{0.9} & \textbf{0.66} & \textbf{0.77} & \underline{0.87} & \textbf{0.82} & \textbf{0.89} & \textbf{0.94} & 0.71 & 0.82 & \textbf{0.95} \\
MPRisk & \textbf{0.76} & \textbf{0.84} & \textbf{0.91} & \textbf{0.66} & \textbf{0.77} & \textbf{0.88} & \underline{0.81} & \textbf{0.89} & \textbf{0.94} & 0.75 & \textbf{0.89} & \textbf{0.95} \\
Rejected-SCF & 0.28 & 0.15 & 0.08 & 0.29 & 0.16 & 0.08 & 0.27 & 0.11 & 0.04 & 0.68 & 0.07 & 0 \\
\bottomrule
\end{tabular}
\end{adjustbox}
\end{table}

\begin{table}[t]
\centering
\scriptsize
\setlength\tabcolsep{2pt}
\caption{Full MPRisk component ablation on open-set text benchmarks. PRR for $F_1$ filtering is reported.}
\label{tab:full_component_ablation_text}
\begin{adjustbox}{width=\textwidth}
\begin{tabular}{lccccccccccccccc}
\toprule
Variant & \multicolumn{3}{c}{Yahoo Answers} & \multicolumn{3}{c}{AGNews} & \multicolumn{3}{c}{DBPedia} & \multicolumn{3}{c}{CLINC150} & \multicolumn{3}{c}{PAN-20-AV} \\
 & $0.1$ & $0.3$ & $0.5$ & $0.1$ & $0.3$ & $0.5$ & $0.1$ & $0.3$ & $0.5$ & $0.1$ & $0.3$ & $0.5$ & $0.1$ & $0.3$ & $0.5$ \\
\midrule
$r_{FA}$ & -1.44 & 0.09 & \underline{0.5} & -1.51 & 0.38 & \underline{0.66} & 0.55 & \textbf{0.92} & \textbf{0.95} & -0.49 & 0.15 & \underline{0.42} & -1.44 & -0.58 & 0.2 \\
$r_{ID}$ & -0.43 & -0.21 & \textbf{0.64} & -0.53 & -0.29 & \textbf{0.76} & -0.04 & -0.05 & -0.12 & -0.45 & 0.01 & 0.13 & -1.41 & -0.45 & 0.58 \\
$r_{FR}$ & \textbf{0.66} & 0.31 & -0.04 & 0.53 & 0.05 & -0.19 & -0.47 & -0.27 & -0.21 & \underline{0.69} & 0.29 & 0.01 & \textbf{0.35} & 0.31 & 0.28 \\
$r_{NS}$ & 0.17 & 0.18 & -0.07 & 0.22 & 0.01 & -0.22 & 0.06 & -0.08 & -0.21 & 0.55 & 0.26 & 0 & 0.25 & 0.22 & 0.28 \\
$r_{FA}+r_{ID}$ & -1.44 & 0.09 & \underline{0.5} & -1.51 & 0.38 & \underline{0.66} & 0.55 & \textbf{0.92} & \textbf{0.95} & -0.4 & 0.2 & 0.33 & -1.39 & -0.38 & \textbf{0.65} \\
$r_{FR}+r_{NS}$ & \textbf{0.66} & 0.31 & -0.04 & \textbf{0.57} & 0.05 & -0.19 & 0.15 & -0.08 & -0.21 & \textbf{0.71} & 0.29 & 0.01 & \textbf{0.35} & 0.31 & 0.28 \\
Ordinary no $r_{NS}$ & -0.15 & \underline{0.43} & 0.49 & 0.02 & \underline{0.47} & 0.51 & 0.65 & \underline{0.83} & \underline{0.79} & 0.14 & 0.4 & 0.41 & -0.14 & \textbf{0.51} & 0.6 \\
Full raw & -0.15 & \underline{0.43} & 0.49 & 0.02 & \underline{0.47} & 0.51 & 0.65 & \underline{0.83} & \underline{0.79} & 0.14 & 0.4 & 0.41 & -0.14 & \textbf{0.51} & 0.6 \\
Tuned no $r_{NS}$ & \textbf{0.66} & 0.18 & \textbf{0.64} & \underline{0.54} & \textbf{0.48} & \textbf{0.76} & \underline{0.69} & \textbf{0.92} & \textbf{0.95} & \textbf{0.71} & \textbf{0.6} & \textbf{0.59} & \underline{0.33} & \underline{0.49} & \underline{0.64} \\
MPRisk & \textbf{0.66} & \textbf{0.46} & \textbf{0.64} & 0.53 & \textbf{0.48} & \textbf{0.76} & \textbf{0.72} & \textbf{0.92} & \textbf{0.95} & \textbf{0.71} & \underline{0.59} & \textbf{0.59} & 0.31 & 0.47 & 0.61 \\
Rejected-SCF & \underline{0.18} & 0.18 & -0.07 & 0.22 & 0.01 & -0.22 & 0.05 & -0.08 & -0.21 & 0.59 & 0.27 & 0 & 0.25 & 0.22 & 0.28 \\
\bottomrule
\end{tabular}
\end{adjustbox}
\end{table}

Three facts emerge.
First, single components are not scores: each decision-conditioned component is identically zero on the complementary decision set, so in isolation it ranks half of the probes arbitrarily.
This is why $r_{\mathrm{FA}}$ alone reaches PRR $-3.54$ at low FPIR (where almost no unknown probes are accepted) yet $0.85$ at high FPIR.
Second, which component dominates is dataset- and operating-point-specific: acceptance risks dominate DBPedia and high-FPIR points, while rejection components dominate Yahoo, CLINC150, and PAN at low FPIR ($r_{\mathrm{FR}}$ alone achieves $0.66$ on Yahoo).
No fixed weighting serves all regimes.
Third, the tuned combination is uniformly near the per-cell maximum over all variants, while the equal-cost combinations inherit the weaknesses of their dominant untuned term.
The addition of $r_{\mathrm{NS}}$ on top of the tuned three-component score gives small consistent gains at specific operating points ($+0.01$ on IJB-C/IJB-B at FPIR $0.2$; $+0.04$/$+0.07$ on VoxBlink; $+0.03$ on DBPedia at FPIR $0.1$; $+0.28$ on Yahoo at FPIR $0.3$); its main value, shown in Sections~\ref{sec:mixedprior}--\ref{sec:errortypes}, is rejection-specific detection and interpretability.

\subsection{Is the mixed prior necessary?}\label{sec:mixedprior}

Tables~\ref{tab:mixed_prior_necessity_bio} and~\ref{tab:mixed_prior_necessity_text} isolate the contribution of modeling the unknown class as a continuum.

\begin{table}[t]
\centering
\scriptsize
\setlength\tabcolsep{2pt}
\caption{Mixed-prior necessity ablation on image and Whale benchmarks. PRR for $F_1$ filtering is reported. The collapsed-unknown variant replaces the continuous unknown component by a single reject class, making $r_{NS}$ unavailable; consequently the first three rows coincide by construction.}
\label{tab:mixed_prior_necessity_bio}
\begin{adjustbox}{width=0.75\textwidth}
\begin{tabular}{lcccccc}
\toprule
Variant & \multicolumn{3}{c}{IJB-C} & \multicolumn{3}{c}{Whale} \\
 & $0.05$ & $0.1$ & $0.2$ & $0.05$ & $0.1$ & $0.2$ \\
\midrule
Collapsed unknown & \underline{0.6} & \underline{0.44} & \underline{0.31} & \underline{0.78} & \underline{0.76} & \underline{0.7} \\
No $r_{NS}$ & \underline{0.6} & \underline{0.44} & \underline{0.31} & \underline{0.78} & \underline{0.76} & \underline{0.7} \\
Equal full & \underline{0.6} & \underline{0.44} & \underline{0.31} & \underline{0.78} & \underline{0.76} & \underline{0.7} \\
$r_{NS}$ & 0.28 & 0.15 & 0.06 & 0.27 & 0.11 & 0.04 \\
$\mathcal{N}_0$ & 0.4 & 0.31 & 0.23 & 0.16 & 0.02 & -0.06 \\
MPRisk & \textbf{0.76} & \textbf{0.83} & \textbf{0.91} & \textbf{0.81} & \textbf{0.88} & \textbf{0.94} \\
\bottomrule
\end{tabular}
\end{adjustbox}
\end{table}

\begin{table}[t]
\centering
\scriptsize
\setlength\tabcolsep{2pt}
\caption{Mixed-prior necessity ablation on text benchmarks. PRR for $F_1$ filtering is reported.}
\label{tab:mixed_prior_necessity_text}
\begin{adjustbox}{width=\textwidth}
\begin{tabular}{lcccccccccccc}
\toprule
Variant & \multicolumn{3}{c}{Yahoo Answers} & \multicolumn{3}{c}{DBPedia} & \multicolumn{3}{c}{CLINC150} & \multicolumn{3}{c}{PAN-20-AV} \\
 & $0.1$ & $0.3$ & $0.5$ & $0.1$ & $0.3$ & $0.5$ & $0.1$ & $0.3$ & $0.5$ & $0.1$ & $0.3$ & $0.5$ \\
\midrule
Collapsed unknown & -0.15 & \underline{0.43} & 0.49 & \underline{0.65} & \underline{0.83} & \underline{0.79} & 0.14 & \underline{0.4} & \underline{0.41} & -0.14 & \textbf{0.51} & \underline{0.6} \\
No $r_{NS}$ & -0.15 & \underline{0.43} & 0.49 & \underline{0.65} & \underline{0.83} & \underline{0.79} & 0.14 & \underline{0.4} & \underline{0.41} & -0.14 & \textbf{0.51} & \underline{0.6} \\
Equal full & -0.15 & \underline{0.43} & 0.49 & \underline{0.65} & \underline{0.83} & \underline{0.79} & 0.14 & \underline{0.4} & \underline{0.41} & -0.14 & \textbf{0.51} & \underline{0.6} \\
$r_{NS}$ & \underline{0.17} & 0.18 & -0.07 & 0.06 & -0.08 & -0.21 & \underline{0.55} & 0.26 & 0 & \underline{0.25} & 0.22 & 0.28 \\
$\mathcal{N}_0$ & 0.16 & 0.39 & \underline{0.55} & 0.01 & 0.14 & 0.31 & -0.03 & -0.16 & -0.26 & 0.02 & 0.05 & 0.22 \\
MPRisk & \textbf{0.66} & \textbf{0.46} & \textbf{0.64} & \textbf{0.72} & \textbf{0.92} & \textbf{0.95} & \textbf{0.71} & \textbf{0.59} & \textbf{0.59} & \textbf{0.37} & \underline{0.48} & \textbf{0.62} \\
\bottomrule
\end{tabular}
\end{adjustbox}
\end{table}

Under a collapsed unknown class the three ordinary risks are unchanged, so the first three rows coincide: the only score the mixed prior adds is $r_{\mathrm{NS}}$.
Two nested comparisons quantify its ingredients.
The raw non-specificity $\calN_0$ alone equals the SCF quality score up to a monotone transform (compare its IJB-C row, $0.40/0.31/0.23$, with SCF in Table~\ref{tab:mprisk_main_prr_bio}) and is weak.
Gating and $P_0$ weighting change its behavior qualitatively --- as a standalone score it becomes a rejection specialist nearly identical to Rejected-SCF --- and its value materializes inside the tuned combination.
Most of the improvement over the equal-weight score comes from tuning the three standard decision risks (Tables~\ref{tab:full_component_ablation_bio}--\ref{tab:full_component_ablation_text}); the non-specificity penalty provides smaller aggregate gains but adds a rejection-specific signal and makes suspicious rejects interpretable.

\subsection{Error-type detection}\label{sec:errortypes}

PRR aggregates all error types into one ranking; Tables~\ref{tab:error_type_detection_bio} and~\ref{tab:error_type_detection_text} disaggregate them.
MPRisk cal is reported alongside MPRisk: monotone calibration preserves AUROC up to ties, so the rows are near-identical.

\begin{table}[t]
\centering
\footnotesize
\setlength\tabcolsep{4pt}
\caption{Error-type detection quality on IJB-C at FPIR $0.05$ and Whale at FPIR $0.1$. AUROC is reported for detecting any OSR error (Any), false acceptance (FA), false rejection (FR), and misidentification (ID).}
\label{tab:error_type_detection_bio}
\begin{tabular}{lcccccccc}
\toprule
 & \multicolumn{4}{c}{IJB-C (FPIR $0.05$)} & \multicolumn{4}{c}{Whale (FPIR $0.1$)} \\
\cmidrule(lr){2-5}\cmidrule(lr){6-9}
Method & Any & FA & FR & ID & Any & FA & FR & ID \\
\midrule
SCF & 0.70 & 0.61 & 0.77 & \underline{0.91} & 0.61 & 0.55 & 0.84 & 0.86 \\
AccScr & 0.87 & 0.91 & \textbf{0.80} & 0.85 & 0.88 & 0.87 & \textbf{0.86} & 0.81 \\
MSP & 0.87 & 0.91 & \textbf{0.80} & 0.88 & 0.88 & 0.88 & \underline{0.85} & 0.88 \\
GalUE & \underline{0.88} & 0.92 & \textbf{0.80} & 0.90 & 0.89 & 0.88 & \textbf{0.86} & \underline{0.90} \\
HolUE & 0.87 & \underline{0.96} & 0.70 & \textbf{0.97} & \underline{0.93} & \underline{0.93} & \underline{0.85} & \underline{0.90} \\
MPRisk raw & 0.81 & 0.80 & \underline{0.79} & 0.84 & 0.89 & 0.88 & \textbf{0.86} & \underline{0.90} \\
MPRisk & \textbf{0.89} & \textbf{0.98} & 0.75 & \textbf{0.97} & \textbf{0.95} & \textbf{0.96} & 0.79 & \textbf{0.94} \\
MPRisk cal & \textbf{0.89} & \textbf{0.98} & 0.75 & \textbf{0.97} & \textbf{0.95} & \textbf{0.96} & 0.79 & \textbf{0.94} \\
\bottomrule
\end{tabular}
\end{table}

\begin{table}[t]
\centering
\footnotesize
\setlength\tabcolsep{4pt}
\caption{Error-type detection quality on Yahoo Answers and PAN-20-AV, both at FPIR $0.1$. AUROC is reported for detecting any OSR error (Any), false acceptance (FA), false rejection (FR), and misidentification (ID); misidentification is absent in the Yahoo Answers protocol at this operating point.}
\label{tab:error_type_detection_text}
\begin{tabular}{lcccccccc}
\toprule
 & \multicolumn{4}{c}{Yahoo Answers (FPIR $0.1$)} & \multicolumn{4}{c}{PAN-20-AV (FPIR $0.1$)} \\
\cmidrule(lr){2-5}\cmidrule(lr){6-9}
Method & Any & FA & FR & ID & Any & FA & FR & ID \\
\midrule
SCF & 0.45 & 0.66 & 0.41 & -- & 0.50 & 0.37 & \underline{0.56} & 0.36 \\
AccScr & 0.59 & \textbf{0.77} & 0.54 & -- & 0.61 & \textbf{0.89} & 0.48 & 0.83 \\
MSP & 0.54 & \underline{0.76} & 0.48 & -- & \underline{0.64} & 0.82 & 0.53 & 0.86 \\
GalUE & 0.58 & \textbf{0.77} & 0.53 & -- & 0.61 & \underline{0.88} & 0.48 & \underline{0.90} \\
HolUE & \textbf{0.89} & 0.58 & \textbf{0.92} & -- & 0.59 & 0.75 & 0.52 & 0.70 \\
MPRisk raw & 0.59 & \textbf{0.77} & 0.53 & -- & 0.61 & \underline{0.88} & 0.48 & \textbf{0.91} \\
MPRisk & \underline{0.79} & 0.20 & \underline{0.90} & -- & \textbf{0.68} & 0.62 & \textbf{0.66} & 0.66 \\
MPRisk cal & \underline{0.79} & 0.27 & 0.88 & -- & \textbf{0.68} & 0.62 & \textbf{0.66} & 0.66 \\
\bottomrule
\end{tabular}
\end{table}

On the image and audio benchmarks, tuned MPRisk is the best overall error detector and the best FA and ID detector (IJB-C: FA $0.98$, ID $0.97$; Whale: FA $0.96$, ID $0.94$), at a moderate cost in FR AUROC relative to threshold-based scores.
On Yahoo Answers, where errors are almost entirely false rejections, optimization for aggregate PRR pushes MPRisk's weights toward $r_{\mathrm{FR}}$/$r_{\mathrm{NS}}$: its FR AUROC reaches $0.90$, close to HolUE's $0.92$, while its FA AUROC drops to $0.20$ because the objective assigns little weight to the rare FA errors.
On PAN-20-AV, MPRisk is the best any-error and FR detector ($0.68$ and $0.66$) while conceding FA and ID to gallery scores; MPRisk raw is the best ID detector there ($0.91$).
These trade-offs stem from compressing four costs into one scalar for one metric, not from the decomposition itself: a deployment with asymmetric error costs can tune $\lambda$ against its own cost matrix, which single-summary scores do not support.

\subsection{Statistical significance}\label{sec:significance}

Tables~\ref{tab:bootstrap_bio} and~\ref{tab:bootstrap_text} report paired bootstrap confidence intervals for PRR differences.

\begin{table}[t]
\centering
\footnotesize
\setlength\tabcolsep{3pt}
\caption{Paired bootstrap confidence intervals for PRR differences on image, Whale, and audio benchmarks. A reported $p(\Delta\leq 0)$ of $0$ indicates that no bootstrap resample produced $\Delta\leq 0$.}
\label{tab:bootstrap_bio}
\begin{adjustbox}{width=\textwidth}
\begin{tabular}{lcccccc}
\toprule
Dataset & FPIR & Comparison & PRR A & PRR B & $\Delta$ PRR [95\% CI] & $p(\Delta\leq 0)$ \\
\midrule
IJB-C & $0.05$ & MPRisk $-$ HolUE & 0.76 & 0.76 & $0.01\,[-0.01,0.03]$ & 0.166 \\
IJB-B & $0.05$ & MPRisk $-$ HolUE & 0.67 & 0.54 & $0.12\,[0.07,0.17]$ & 0 \\
Whale & $0.1$ & MPRisk $-$ HolUE & 0.88 & 0.82 & $0.06\,[0.04,0.08]$ & 0 \\
VB-Eval-L-5 & $0.05$ & MPRisk $-$ HolUE & 0.89 & 0.81 & $0.08\,[0.06,0.1]$ & 0 \\
IJB-C & $0.05$ & MPRisk cal $-$ HolUE & 0.76 & 0.76 & $0.01\,[-0.01,0.03]$ & 0.178 \\
IJB-B & $0.05$ & MPRisk cal $-$ HolUE & 0.67 & 0.54 & $0.12\,[0.07,0.17]$ & 0 \\
Whale & $0.1$ & MPRisk cal $-$ HolUE & 0.88 & 0.82 & $0.06\,[0.04,0.08]$ & 0 \\
VB-Eval-L-5 & $0.05$ & MPRisk cal $-$ HolUE & 0.89 & 0.81 & $0.08\,[0.06,0.1]$ & 0 \\
IJB-C & $0.05$ & MPRisk $-$ MPRisk raw & 0.76 & 0.6 & $0.16\,[0.13,0.19]$ & 0 \\
IJB-C & $0.05$ & MPRisk $-$ MPRisk no $r_{NS}$ & 0.76 & 0.6 & $0.16\,[0.13,0.19]$ & 0 \\
Whale & $0.1$ & MPRisk $-$ MPRisk raw & 0.88 & 0.76 & $0.12\,[0.1,0.14]$ & 0 \\
Whale & $0.1$ & MPRisk $-$ MPRisk no $r_{NS}$ & 0.88 & 0.76 & $0.12\,[0.1,0.14]$ & 0 \\
\bottomrule
\end{tabular}
\end{adjustbox}
\end{table}

\begin{table}[t]
\centering
\footnotesize
\setlength\tabcolsep{3pt}
\caption{Paired bootstrap confidence intervals for PRR differences on text benchmarks. A reported $p(\Delta\leq 0)$ of $0$ indicates that no bootstrap resample produced $\Delta\leq 0$.}
\label{tab:bootstrap_text}
\begin{adjustbox}{width=\textwidth}
\begin{tabular}{lcccccc}
\toprule
Dataset & FPIR & Comparison & PRR A & PRR B & $\Delta$ PRR [95\% CI] & $p(\Delta\leq 0)$ \\
\midrule
Yahoo Answers & $0.1$ & MPRisk $-$ HolUE & 0.66 & 0.77 & $-0.11\,[-0.13,-0.09]$ & 1 \\
AGNews & $0.1$ & MPRisk $-$ HolUE & 0.54 & 0.47 & $0.07\,[-0.02,0.17]$ & 0.073 \\
DBPedia & $0.1$ & MPRisk $-$ HolUE & 0.72 & 0.52 & $0.19\,[0.15,0.25]$ & 0 \\
CLINC150 & $0.1$ & MPRisk $-$ HolUE & 0.71 & 0.69 & $0.02\,[0,0.04]$ & 0.046 \\
PAN-20-AV & $0.1$ & MPRisk $-$ HolUE & 0.37 & 0.18 & $0.18\,[-0.05,0.4]$ & 0.054 \\
Yahoo Answers & $0.1$ & MPRisk cal $-$ HolUE & 0.66 & 0.77 & $-0.11\,[-0.13,-0.09]$ & 1 \\
AGNews & $0.1$ & MPRisk cal $-$ HolUE & 0.54 & 0.47 & $0.07\,[-0.02,0.16]$ & 0.063 \\
DBPedia & $0.1$ & MPRisk cal $-$ HolUE & 0.72 & 0.52 & $0.19\,[0.14,0.25]$ & 0 \\
CLINC150 & $0.1$ & MPRisk cal $-$ HolUE & 0.69 & 0.69 & $-0.01\,[-0.03,0.02]$ & 0.558 \\
PAN-20-AV & $0.1$ & MPRisk cal $-$ HolUE & 0.37 & 0.18 & $0.18\,[-0.04,0.39]$ & 0.062 \\
Yahoo Answers & $0.1$ & MPRisk $-$ MPRisk raw & 0.66 & -0.15 & $0.8\,[0.77,0.86]$ & 0 \\
Yahoo Answers & $0.1$ & MPRisk $-$ MPRisk no $r_{NS}$ & 0.66 & -0.15 & $0.8\,[0.77,0.87]$ & 0 \\
PAN-20-AV & $0.1$ & MPRisk $-$ MPRisk raw & 0.37 & -0.14 & $0.51\,[0.2,0.78]$ & 0 \\
PAN-20-AV & $0.1$ & MPRisk $-$ MPRisk no $r_{NS}$ & 0.37 & -0.14 & $0.51\,[0.21,0.8]$ & 0.001 \\
\bottomrule
\end{tabular}
\end{adjustbox}
\end{table}

On the image and audio benchmarks, the 95\% intervals for the gain over HolUE exclude zero on IJB-B ($\Delta=0.12$, $[0.07,0.17]$), Whale ($0.06$, $[0.04,0.08]$), and VoxBlink ($0.08$, $[0.06,0.10]$); IJB-C at FPIR $0.05$ is a statistical tie.
On text, the intervals exclude zero on DBPedia ($0.19$, $[0.15,0.25]$) and CLINC150 ($p=0.046$), include zero on AGNews ($p=0.073$) and PAN-20-AV ($p=0.054$; the wide interval reflects the small PAN test set), and MPRisk is significantly worse on Yahoo Answers ($-0.11$).
The gain of tuned MPRisk over MPRisk raw exceeds zero in every tested comparison, and monotone calibration leaves all conclusions unchanged (MPRisk cal rows).

\subsection{Calibration}\label{sec:calibration}

\begin{figure}[t]
\centering
\includegraphics[width=0.48\linewidth]{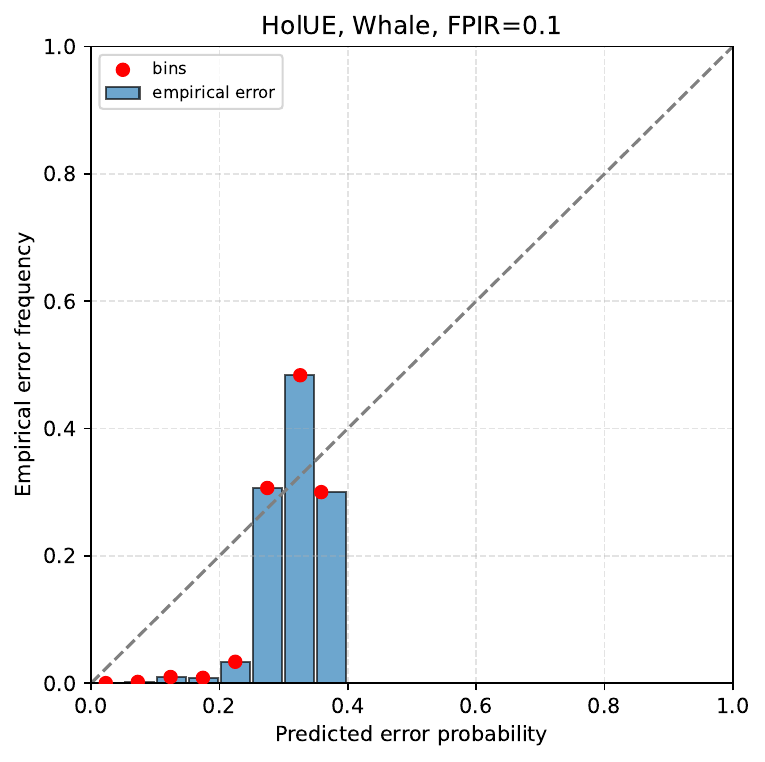}\hfill
\includegraphics[width=0.48\linewidth]{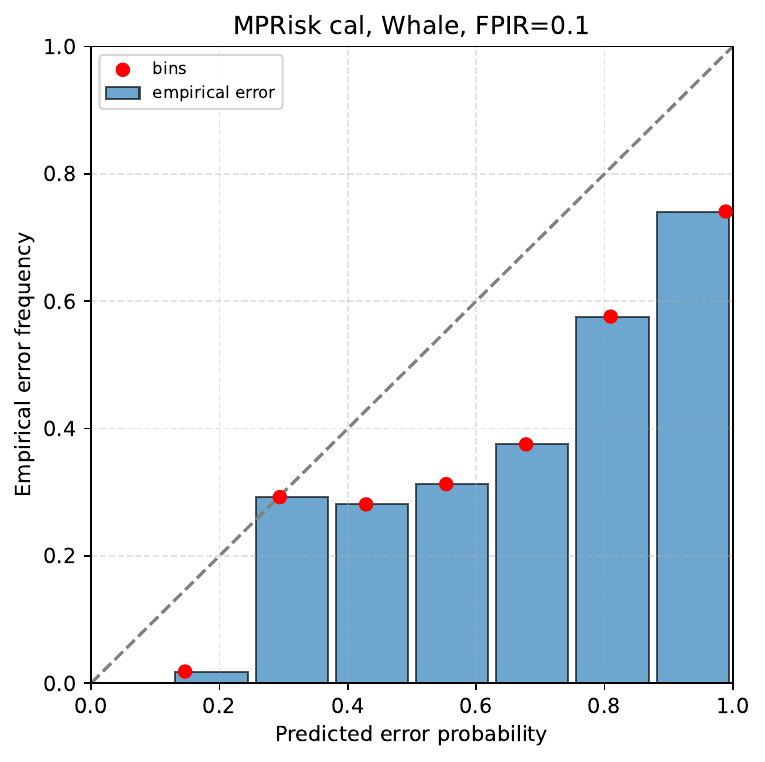}\\[4pt]
\includegraphics[width=0.48\linewidth]{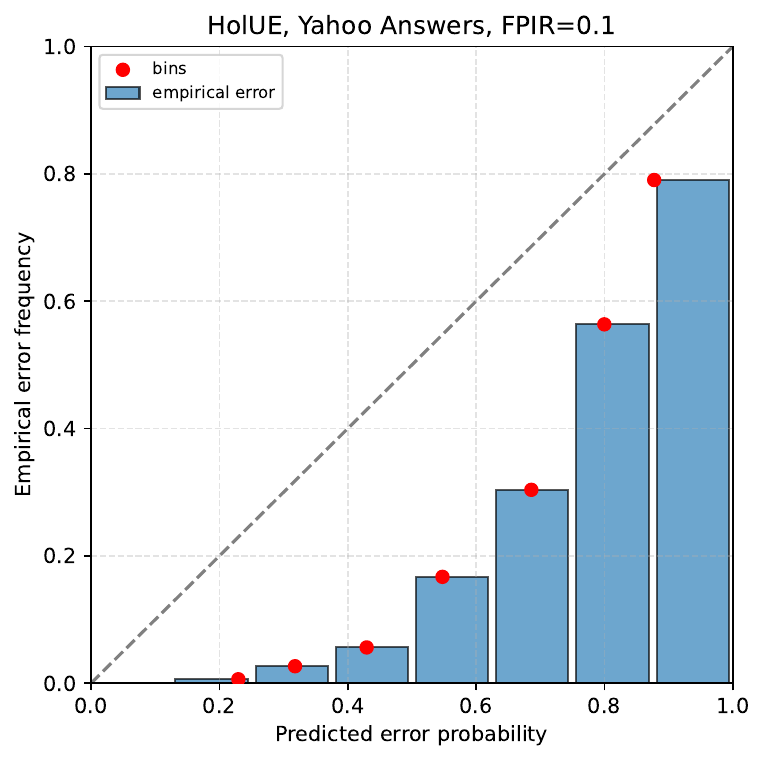}\hfill
\includegraphics[width=0.48\linewidth]{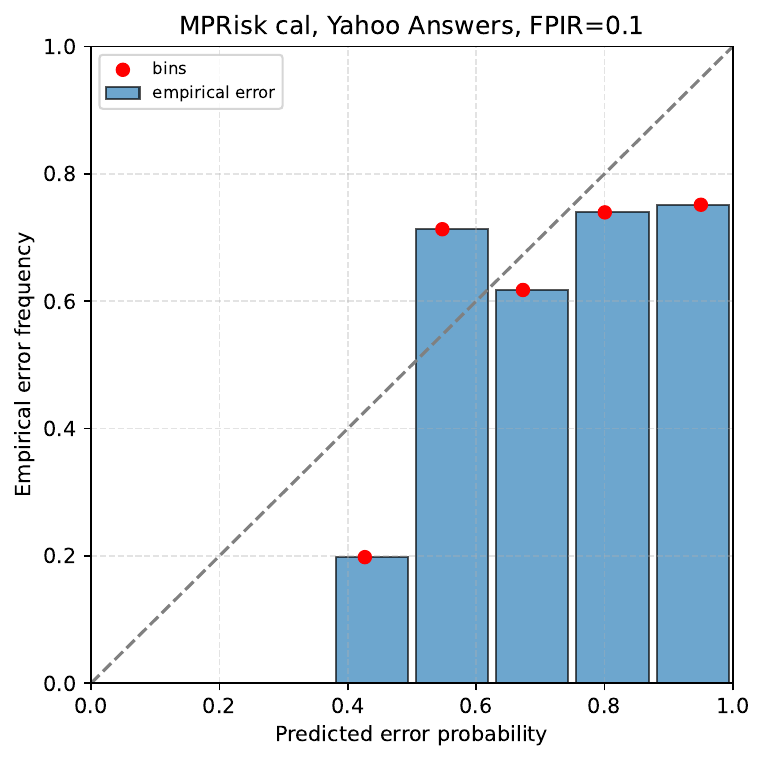}
\caption{Reliability diagrams for calibrated error probabilities of HolUE and MPRisk cal on Whale at FPIR $0.1$ (top) and Yahoo Answers at FPIR $0.1$ (bottom).
Predicted error probability (binned) is plotted against empirical error frequency per bin; the diagonal denotes perfect calibration.
}
\label{fig:calibration}
\end{figure}

Table~\ref{tab:reliability} evaluates whether the scores double as error-probability estimates; Figure~\ref{fig:calibration} shows reliability diagrams.
Both compared quantities are validation-calibrated --- HolUE by its calibration model, MPRisk cal by the monotone calibrator --- so this comparison is also symmetric.

\begin{table}[t]
\centering
\footnotesize
\setlength\tabcolsep{3pt}
\caption{Calibration quality (ECE, $\downarrow$) on image, Whale, audio, and text benchmarks. The FPIR operating point is indicated below each dataset name.}
\label{tab:reliability}
\begin{adjustbox}{width=\textwidth}
\begin{tabular}{lccccccc}
\toprule
Method & IJB-C & Whale & VB-Eval-L-5 & Yahoo Answers & DBPedia & CLINC150 & PAN-20-AV \\
 & $0.05$ & $0.1$ & $0.05$ & $0.1$ & $0.1$ & $0.1$ & $0.1$ \\
\midrule
HolUE & \underline{0.049} & \textbf{0.125} & \textbf{0.111} & \underline{0.314} & \textbf{0.407} & \textbf{0.240} & \underline{0.207} \\
MPRisk cal & \textbf{0.020} & \underline{0.137} & \underline{0.165} & \textbf{0.228} & \underline{0.416} & \underline{0.254} & \textbf{0.192} \\
\bottomrule
\end{tabular}
\end{adjustbox}
\end{table}

Calibration results are mixed: MPRisk cal has lower ECE on IJB-C ($0.020$ vs.\ $0.049$), Yahoo Answers, and PAN-20-AV, while HolUE is lower on Whale, VoxBlink, DBPedia, and CLINC150 by smaller margins.
Since calibration does not change the selective-recognition ranking (Tables~\ref{tab:bootstrap_bio}--\ref{tab:bootstrap_text}), the calibrated variant can be used whenever an absolute error-probability readout is needed at no ranking cost.

\subsection{Runtime}\label{sec:runtime}

Table~\ref{tab:runtime} reports online per-probe cost, excluding the shared backbone forward pass.

\begin{table}[t]
\centering
\footnotesize
\setlength\tabcolsep{3pt}
\caption{Runtime overhead on image, Whale, audio, and text benchmarks. Online time in milliseconds per probe is reported (backbone forward pass excluded); the FPIR operating point is indicated below each dataset name.}
\label{tab:runtime}
\begin{adjustbox}{width=\textwidth}
\begin{tabular}{lccccccc}
\toprule
Method & IJB-C & Whale & VB-Eval-L-5 & Yahoo Answers & DBPedia & CLINC150 & PAN-20-AV \\
 & $0.05$ & $0.1$ & $0.05$ & $0.1$ & $0.1$ & $0.1$ & $0.1$ \\
\midrule
HolUE & 2.08 & 3.03 & 2.95 & 3.68 & 2.03 & 4.82 & 132.43 \\
MPRisk raw & 0.67 & 1.35 & 1.55 & 0.09 & 0.05 & 0.14 & 0.45 \\
MPRisk & 1.89 & 3.13 & 2.33 & 2.56 & 1.41 & 1.84 & 31.17 \\
MPRisk cal & 2.07 & 3.08 & 2.38 & 2.55 & 1.54 & 2.04 & 32.14 \\
\bottomrule
\end{tabular}
\end{adjustbox}
\end{table}

MPRisk raw costs $0.05$--$1.6$~ms per probe: its inputs are the posterior probabilities already computed for the decision plus one analytic formula.
Tuned MPRisk and MPRisk cal are comparable to or cheaper than HolUE on all benchmarks, and over four times cheaper on PAN-20-AV ($31$--$32$ vs.\ $132$~ms), where the dynamic per-query gallery makes the KL$_2$ computation of HolUE expensive.

\section{DISCUSSION}\label{sec:discussion}

\paragraph{What the experiments establish.}
HolUE and MPRisk use the same posterior and validation split; they differ in the features and the final scoring model.
The KL components contain useful information, but their ordering is not generally consistent with decision risk, so their usefulness depends on the learned nonlinear calibration: the same features under linear tuning produce negative PRR on several benchmarks.
The decision-conditioned components, in contrast, combine effectively under a four-weight linear rule, generalize stably from validation to test, and outperform the calibrated KL score at 25 of 27 image/audio operating points and on most text operating points.
Overall: gains whose bootstrap intervals exclude zero on five of nine benchmarks, ties or inconclusive differences on three, and one clear HolUE win (Yahoo Answers).

\paragraph{When KL features remain useful.}
The Yahoo result is diagnostic.
Its error mass is almost purely false rejections of ambiguous multi-topic questions, for which the quality-driven $\mathrm{KL}_2$ feature under HolUE's calibration is a near-ideal single signal.
MPRisk's scalar objective gains FR sensitivity (FR AUROC $0.90$) at the expense of FA sensitivity (FA AUROC $0.20$).
Two remedies follow from our results: tune $\lambda$ against the deployment's own cost matrix, or use the hybrid, which recovers most of HolUE's Yahoo performance while preserving MPRisk's gains elsewhere.

\paragraph{The role of the mixed prior.}
The continuous unknown component contributes on three levels: small consistent PRR gains at specific operating points; a rejection-specific detection signal unavailable to a collapsed reject class; and an explanation of \emph{why} a rejection is suspicious --- high unknown mass with no specific unknown hypothesis --- which single-scalar scores cannot articulate.

\paragraph{Capacity and validation size.}
Logistic regression on the full $14$-dimensional feature set exceeds tuned MPRisk only modestly and not uniformly; the generic $14$-input MLP varies substantially across datasets, whereas HolUE's two-input, task-objective MLP is stable.
On moderate validation sets, performance tracks how tightly the hypothesis class matches the task rather than raw capacity; MPRisk's four-weight rule over semantically fixed components is the lowest-capacity option in this family and achieves comparable performance.
When abundant validation errors are available, the hybrid or a logistic detector on the joint features is a reasonable recipe; when validation data are limited, tuned MPRisk is a lower-risk choice and MPRisk raw a tuning-free fallback.

\paragraph{Limitations.}
Tuned MPRisk depends on a validation split matched to the test operating point, and the scalar PRR objective induces error-type trade-offs (Section~\ref{sec:errortypes}); multi-objective or cost-matrix tuning is a direct extension.
The method inherits the assumptions of the Bayesian posterior and the quality of the SCF embedding distribution: the mean-embedding approximation bounds posterior fidelity in extremely low-$\kappa$ regimes, and the interpretation of non-specificity as evidence of false rejection has been validated only within the tested models.
Robustness under gallery shift and validation/test cost mismatch was not evaluated.
Finally, calibration results are mixed, and both methods over-predict error probability in the tails on text.

\section{CONCLUSION}\label{sec:conclusion}

We proposed MPRisk, a mixed-prior posterior decision-risk score for uncertainty estimation in open-set recognition.
Instead of summarizing the gallery-aware Bayesian posterior by information gain and calibrating the result with a supervised nonlinear model, MPRisk directly scores the error events associated with the selected decision --- false acceptance, misidentification, and false rejection --- and adds a closed-form non-specificity penalty that flags confident but poorly supported rejections, enabled by modeling unknown identities as a continuous component.
Four nonnegative validation-tuned weights suffice for ranking.

Across nine image, audio, and text benchmarks under matched validation budgets, MPRisk is best or tied-best at all image and audio operating points and on most text operating points, with bootstrap-confirmed gains over HolUE on five benchmarks, at comparable or lower runtime.
Linear fusion of the KL features performs poorly, whereas the MPRisk components remain competitive with supervised error detectors; combining the two scores gives further gains in several settings.
These results suggest that decision-conditioned posterior risks provide a simple basis for selective OSR, while KL features supply complementary information when sufficient validation data are available.
Future work includes cost-matrix tuning of the weights and transferring decision-conditioned risk scoring to selective generation and hallucination detection in large language models~\cite{toha}.

\bibliographystyle{unsrt}
\bibliography{references}

\end{document}